\documentclass[11pt]{article}

\usepackage[final]{acl}
\usepackage{amsmath}

\usepackage{times}
\usepackage{latexsym}

\usepackage[T1]{fontenc}

\usepackage[utf8]{inputenc}

\usepackage{microtype}

\usepackage{inconsolata}

\usepackage{graphicx}
\usepackage[table]{xcolor}
\usepackage{booktabs}
\usepackage{array}
\usepackage{amsmath,amssymb}
\usepackage{algorithm}
\usepackage{algpseudocode}

\usepackage{amsmath,amssymb}
\usepackage{amsthm}
\usepackage{multirow}

\theoremstyle{plain}
\newtheorem{theorem}{Theorem}[section]

\newtheorem{lemma}[theorem]{Lemma}

\theoremstyle{definition}

\theoremstyle{remark}
\newtheorem{remark}[theorem]{Remark}

\newcommand{\modelrow}[2]{%
  \midrule
  \rowcolor{gray!10}
  \multicolumn{#1}{>{\columncolor{gray!10}[\tabcolsep][\tabcolsep]}c}{\textbf{#2}} \\
  \midrule
}

\newcommand{\oursrule}[1]{%
  \cmidrule(lr){1-#1}
}

\title{
Towards Understanding Pause Token Fine-Tuning Dynamics: \\
A Mode Retention Perspective

}

\author{
  \textbf{Jaehyeon Kim\textsuperscript{1,*}} \quad
  \textbf{Suhwan Kim\textsuperscript{2,*}} \quad
  \textbf{Nakyung Lee\textsuperscript{2}} \quad
  \textbf{Yeongoon Kim\textsuperscript{2}} \\
  \textbf{Jimin Seo\textsuperscript{2}} \quad
  \textbf{Giho Lee\textsuperscript{1}} \quad
  \textbf{Jungwoo Lee\textsuperscript{1,2,\textdagger}} \\
  \textsuperscript{1}HodooAI Lab, Seoul, South Korea \\
  \textsuperscript{2}Seoul National University, Seoul, South Korea \\
  \small{\textsuperscript{*}Equal contribution. \textsuperscript{\textdagger}Corresponding author.} \\
  \small{\textbf{Correspondence:} \href{mailto:junglee@snu.ac.kr}{\texttt{junglee@snu.ac.kr}}}
}

\begin{document}
\maketitle
\begin{abstract}

Pause-token methods improve LLM reasoning by inserting special
tokens into sequences. Prior work explains these gains through computational expressivity. However, there is relatively little investigation into the training dynamics of pause tokens. We explore how pause tokens reshape the training dynamics of fine-tuning. Two controlled pilots expose distinct asymmetries. On a synthetic continual-learning task, masked pauses overwrite a previously-learned distribution roughly 4× less at matched final adaptation (H1, \textbf{mode retention}); on a synthetic math-reasoning probe, the boundary-adjacent token comes to encode substantially more downstream-step information (H2, \textbf{non-myopic compression}). We formalize a training rule consistent with both — Masked Boundary Pause (MBP), pause tokens placed at reasoning-step boundaries with their loss masked. Across 1B-8B Qwen and Llama models, MBP consistently improves reasoning, achieving gains of up to 6 points on math and 2.5 points on code, while preserving general language understanding abilities. We further demonstrate that this mode-preserving strategy extend gains to GRPO. These results recast pause tokens as a training-dynamics intervention on the retention–adaptation trade-off, rather than merely an inference-time computation device.

\end{abstract}

\section{Introduction}

\begin{figure*}
    \centering
    \includegraphics[width=\linewidth]{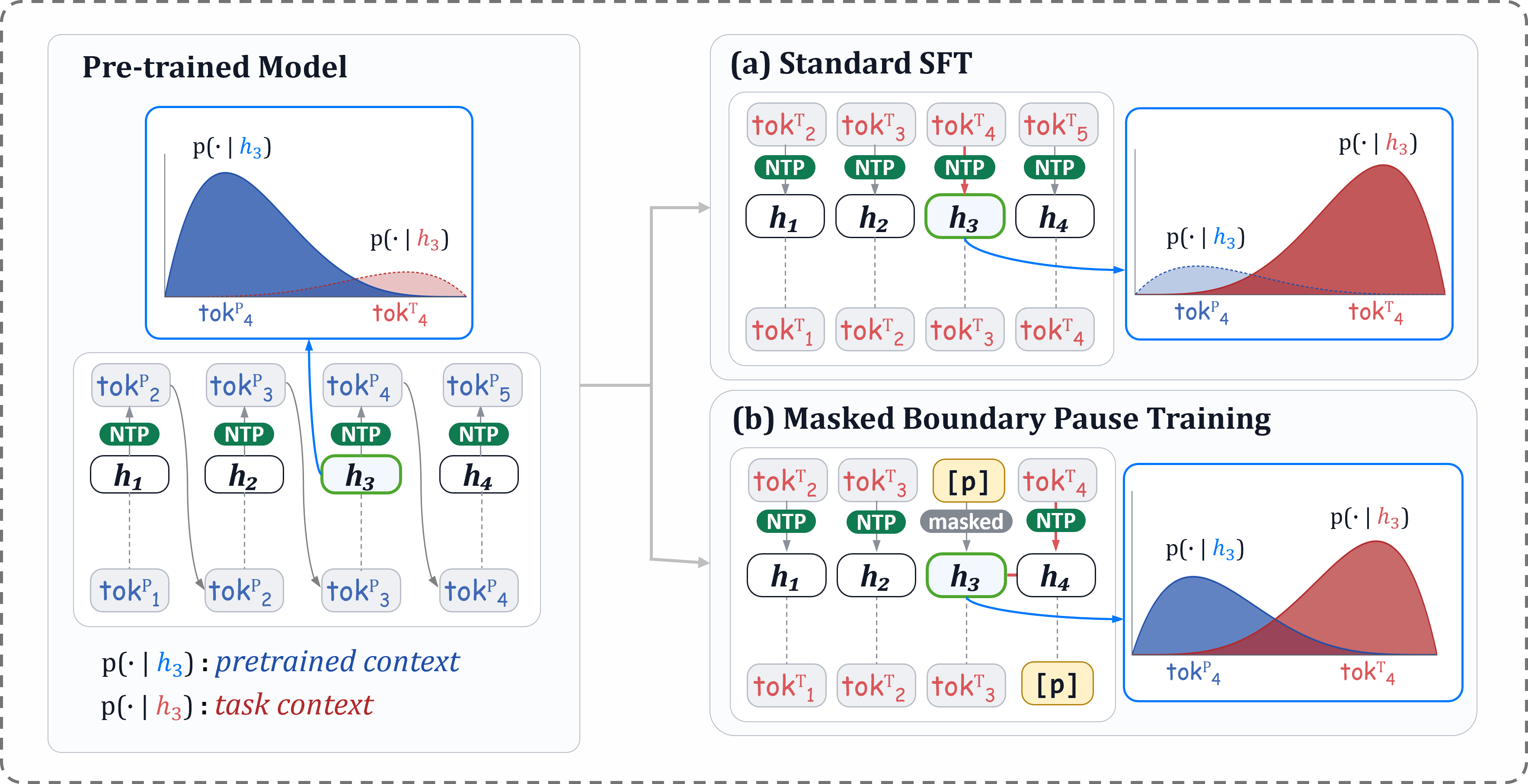}
    \vspace{-15pt}
    \caption{Comparison between standard fine-tuning and masked boundary pause fine-tuning. Standard SFT applies next-token prediction (NTP) loss at every task-token position, allowing gradients from the target token to flow directly into the preceding context representation. This can strongly adapt the model to the task-data distribution, but may also erode the rich predictive distribution learned during pre-training. Masked boundary pause fine-tuning (MBP) instead inserts pause tokens at selected boundaries and masks their loss, thereby blocking the direct gradient path into the boundary context state. The model receives update signals only indirectly through neighboring token predictions, enabling task adaptation while better preserving the original pre-training distribution.}
    \label{fig:main_fig_1}
    \vspace{-10pt}
\end{figure*}

Recent progress in large language models (LLMs) has produced a wide
range of training-time and test-time methods for improving reasoning:
chain-of-thought prompting, test-time scaling, and reinforcement
learning for reasoning achieve strong performance on complex tasks~\cite{cot,scaling,selfconsistency}. However, these methods
incur substantial costs in generation length, inference latency,
sampling budget, or training compute. \emph{Pause-} and
\emph{filler-token} methods stand out within this landscape~\cite{goya,memtrans}. They improve performance through a small and architecture-preserving change by inserting special tokens into sequences without modifying the model or the optimizer.

Pause-token gains have been attributed to forward-pass computation and expressivity~\cite{pfau2024let,pausetoken}, focal-loss amplification at hard tokens~\cite{learntoinsert}, and structural anchoring at sentence
boundaries~\cite{thinksentence}. These explanations center on effects at the inserted positions themselves. Yet much less is known about how pause training shapes the pretrained distribution from which fine-tuning starts, even though this distribution is central to modern adaptation pipelines that increasingly warm-start from existing checkpoints~\cite{gururangan2020don,roziere2023code,azerbayev2023llemma}.

We argue that this is precisely where an underexamined role of pause tokens is most consequential. Pause tokens also participate in the forward pass during fine-tuning, where they alter the context under which target tokens are predicted and reshape how gradient flows through the surrounding positions.~\cite{wu2024do, pfau2024let} These training-dynamics effects are present even when pause tokens are absent at inference and never appear as supervised targets, and they directly modulate the retention--adaptation trade-off central to fine-tuning. In a controlled continual-learning setting, we observe that loss-masked pauses reduce overwriting of a previously learned distribution by roughly $4\times$ at matched final adaptation, revealing a separation that neither inference-time nor low-confidence framings predict.

We investigate this through two controlled pilots that yield distinct
empirical phenomena; a \emph{retention asymmetry} on synthetic
continual learning and an \emph{information-packing asymmetry} on
iGSM~\cite{allenzhu2024igsm} probing, which we elevate to working hypotheses H1 (mode retention)
and H2 (non-myopic compression). 
We then formalize a training rule consistent with both hypotheses, \textbf{Masked Boundary Pause} (MBP)\footnote{Code is available at
\url{https://github.com/jkpizza/emnlp2026-pause-token-dynamics}.}, and test the hypotheses' scaled predictions across 1B--8B fine-tuning and GRPO. The headline finding is that \emph{boundary placement and loss masking are jointly necessary} for the gains, with specific results reported in \S\ref{sec:method}--\S\ref{sec:grpo_results}.

In summary, our contributions are:
\begin{itemize}
\item Through controlled pilots, we identify two hypotheses: (H1)
mode retention, and (H2) non-myopic compression.


\item Motivated by these, we propose \textbf{Masked Boundary Pause
(MBP)} fine-tuning, where pause
tokens are inserted at sentence or reasoning-step boundaries while masking
their cross-entropy loss.

\item Across 1B--8B models, MBP improves math and code reasoning while preserving general capability; gains extend to GRPO. A four-way placement ablation shows that boundary placement and loss masking are jointly necessary.
\end{itemize}


\section{Preliminaries}
\label{sec:prelims}

We formalize masked-pause fine-tuning under a single notation that
subsequent sections re-use, and locate prior pause-token methods
within the same framework.

\subsection{Pause-token training}
\paragraph{Formulation.}
Let $(x, y)$ be a (prompt, response) pair drawn from a fine-tuning
dataset $\mathcal{D}$, with response $y = (y_1, \dots, y_N)$. Standard
supervised fine-tuning (SFT) minimizes the per-token conditional
cross-entropy of $y$ given $x$; we recover it as the
$g(x, y) = \emptyset$ case of the formulation below.

A pause-token training method is parameterized by a
\emph{position selector} $g$ that returns insertion indices
$g(x, y) = \{i_1 < i_2 < \cdots < i_M\} \subseteq \{0, \dots, N\}$.
With the conventions $i_0 = 0$ and $i_{M+1} = N$, these indices
partition $y$ into $M+1$ contiguous segments
\begin{equation}
u_n = (y_{i_{n-1}+1}, \ldots, y_{i_n}), \quad n = 1, \dots, M+1,
\label{eq:segments}
\end{equation}
with $y = u_1 \parallel u_2 \parallel \cdots \parallel u_{M+1}$. A
distinguished pause token $\tau$ is inserted between adjacent
segments, yielding the pause-augmented sequence
\begin{equation}
\tilde y = u_1 \parallel \tau \parallel u_2 \parallel \tau \parallel
\cdots \parallel \tau \parallel u_{M+1}
\label{eq:augmented}
\end{equation}
of length $N + M$. Let
$S_\tau = \{k : \tilde y_{k+1} = \tau\}$ be the set of prediction
positions whose target is a pause token. Following prior work~\cite{goya,learntoinsert}, we
exclude $S_\tau$ from the loss: pause tokens appear in the
autoregressive context but never as supervised outputs. The
per-example training loss is
\begin{equation}
  \mathcal{L}^{g}(\theta)
  = -\frac{1}{N}\sum_{k \notin S_\tau}
  \log p_\theta\bigl(\tilde y_{k+1} \mid x, \tilde y_{1:k}\bigr),
  \label{eq:pause_loss}
\end{equation}
averaged over $(x, y) \sim \mathcal{D}$. Setting $g(x, y) = \emptyset$ recovers standard SFT.


\paragraph{Relation to existing work.}
Prior work differs in the position selector $g$:
\textsc{Append}~\cite{goya,pfau2024let},
\textsc{DIT}~\cite{learntoinsert}, and \textsc{Boundary}
at sentence endings~\cite{thinksentence}. We additionally test
\textsc{Random} placement as an unstructured control. Of these,
\citet{goya} and \citet{learntoinsert} exclude pause positions from
the loss as in Eq.~\ref{eq:pause_loss}, whereas \citet{thinksentence}
trains pauses as predicted targets. Our method,
Masked Boundary Pause Training (\S\ref{sec:method}), takes
$g = \textsc{Boundary}$ with $S_\tau$ masked, and
Section~\ref{sec:sft_setup} compares the four placements.

\subsection{GRPO}
\paragraph{Formulation.}
Group Relative Policy Optimization (GRPO)~\cite{deepseekmath}
optimizes a policy $\pi_\theta$ on prompts $q \in \mathcal{Q}$ by
sampling $G$ responses $\{y_i\}_{i=1}^G \sim
\pi_{\theta_{\text{old}}}(\cdot \mid q)$ from the previous policy,
scoring each with a reward $r_i = R(q, y_i)$, and computing the
group-relative advantage $A_i = (r_i - \mu_r) / (\sigma_r +
\epsilon)$ where $\mu_r, \sigma_r$ are the group's mean and standard
deviation of rewards. With per-token policy ratio
$\rho_{i,t}(\theta) = \pi_\theta(y_{i,t} \mid q, y_{i,<t}) /
\pi_{\theta_{\text{old}}}(y_{i,t} \mid q, y_{i,<t})$, GRPO maximizes

\begin{equation}
\begin{aligned}
\mathcal{J}_{\text{GRPO}}(\theta)
  &= \mathbb{E}\!\left[
       \frac{1}{G}\sum_{i,t} c_{i,t}(\theta)
       - \beta D_{\text{KL}}
     \right], \\[4pt]
c_{i,t}(\theta)
  &= \min\!\bigl(\rho_{i,t}\, A_i,\, \\
  &\qquad\quad \mathrm{clip}(\rho_{i,t},\, 1{-}\epsilon,\, 1{+}\epsilon)\, A_i \bigr).
\end{aligned}
\label{eq:grpo}
\end{equation}

\noindent where $D_{\text{KL}}$ is a per-token KL penalty against a
reference policy.

\section{Pilot Experiments}
\label{sec:pilots}

The introduction posed a mechanistic question: what is masked-pause
fine-tuning actually doing to a pretrained model's representations? We
approach this in two controlled studies. The first asks
whether masked pauses change \emph{what gets overwritten} during
adaptation (\S\ref{sec:pilot_mlc}). The second asks \emph{what gets
encoded} at the boundary-adjacent token (\S\ref{sec:probing_pilot}).
Each pilot yields a distinct empirical asymmetry that we crystallize
into a working hypothesis.

\subsection{Modular continual learning: a retention asymmetry}
\label{sec:pilot_mlc}

Our first pilot asks whether masked pauses change what gets overwritten during adaptation. We use the simplest setting that can support a clean before/after comparison, a synthetic continual-learning task with two disjoint regimes. Detailed experimental settings are provided in Appendix~\ref{app:pilot}; Appendix~\ref{app:pilot_mlc_multistage} extends the pilot to a four-regime chain.


\paragraph{Setup.}
Sequences are generated by a \emph{Modular Linear Chain} (MLC),
\begin{equation}
x_{t+1} = (a_t\, x_t + b_t) \bmod p ,
\label{eq:mlc}
\end{equation}
where a \emph{regime} is defined by a length-$L$ coefficient sequence
$\{(a_t, b_t)\}_{t=1}^L$. We sample two disjoint coefficient sets to
define regimes $D_1$ and $D_2$, and train a small GPT-style
Transformer sequentially: Phase~1 trains on $D_1$ until accuracy
exceeds $0.99$; Phase~2 forks from a single Phase-1 checkpoint and
continues on $D_2$ under two matched conditions sharing architecture,
optimizer, batch size, learning rate, and update budget:
\begin{itemize}
\setlength\itemsep{0pt}
\item \textsc{No pause}: standard next-token prediction on raw $D_2$
  sequences.
\item \textsc{Pause}: one pause token inserted after every content
  token; pause-target positions are excluded from the loss.
\end{itemize}

\begin{figure}
    \centering
    \includegraphics[width=1\linewidth]{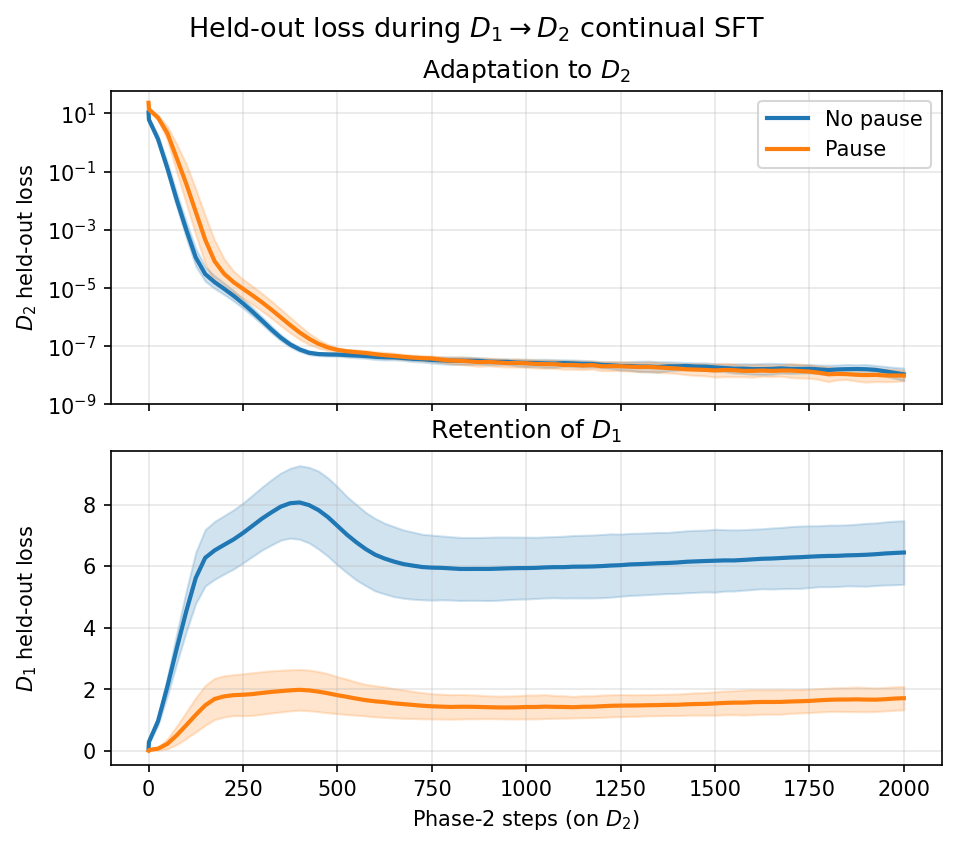}
    \vspace{-20pt}
    \caption{Phase-2 dynamics from a $D_1$-pretrained backbone ($n{=}8$ seeds, $\pm 1\sigma$). \textbf{Top:} $D_2$ adaptation
    (held-out loss, log scale). \textbf{Bottom:} $D_1$ retention (held-out loss, linear scale). Both conditions reach the same
    $D_2$ floor; \textsc{Pause} retains $D_1$ $\sim 4\times$ better than \textsc{No pause}.}
    \label{fig:pilot_mlc}
\end{figure}


\paragraph{Results.}
Figure~\ref{fig:pilot_mlc} shows the Phase-2 loss curves on both regimes. Both conditions adapt to $D_2$ to the same $\sim 10^{-7}$ floor, with \textsc{No pause} marginally faster in descent
but indistinguishable at convergence. The conditions diverge sharply
in how much they overwrite $D_1$: \textsc{No pause} plateaus near
$6$ nats of held-out $D_1$ loss while \textsc{Pause} plateaus near
$1.5$ --- a $\sim 4\times$ retention gap, consistent across seeds. The
gap occurs at \emph{matched} $D_2$ fit and is not explained by
tokens-processed per step.

\paragraph{Hypothesis.}
Masked-pause training matches the new regime's final fit while
preserving the previous regime substantially better:
\begin{quote}
\textbf{H1 (mode retention).} \emph{Masked-pause training reduces
overwriting of previously-learned distributions during adaptation,
relative to standard fine-tuning at matched final performance.}
\end{quote}


\subsection{iGSM step probing: an information-packing asymmetry}
\label{sec:probing_pilot}

The MLC pilot shows masked pauses change adaptation dynamics, but
not what the representation absorbs. To localize the mechanism, we
probe what the \emph{boundary-adjacent token} --- the period
immediately preceding an inserted \texttt{[PAUSE]} --- encodes after
training. We use iGSM~\cite{allenzhu2024igsm}, a synthetic mathematical reasoning dataset where each reasoning
step concludes with a deterministic integer answer, providing the
clean per-step probe targets this analysis requires.


\paragraph{Setup.}
We supervise-fine-tune Qwen3-0.6B-Base~\cite{yang2025qwen3} on 3K iGSM problems under two
conditions --- \textsc{No pause} and \textsc{Pause} (one
\texttt{[PAUSE]} inserted at each step boundary, loss masked) --- and
compare against the pretrained \textsc{Base} checkpoint ($n{=}3$
seeds each). For each held-out problem we extract the final-layer
hidden state at the boundary-adjacent period of every reasoning step
and fit two linear probes: one for the integer answer of the
\emph{current} step, one for the \emph{next} step. Both fine-tuned conditions reach comparable held-out task accuracy, so the probe compares representations at matched task fit.

\begin{figure}
    \centering
    \includegraphics[width=1.0\linewidth]{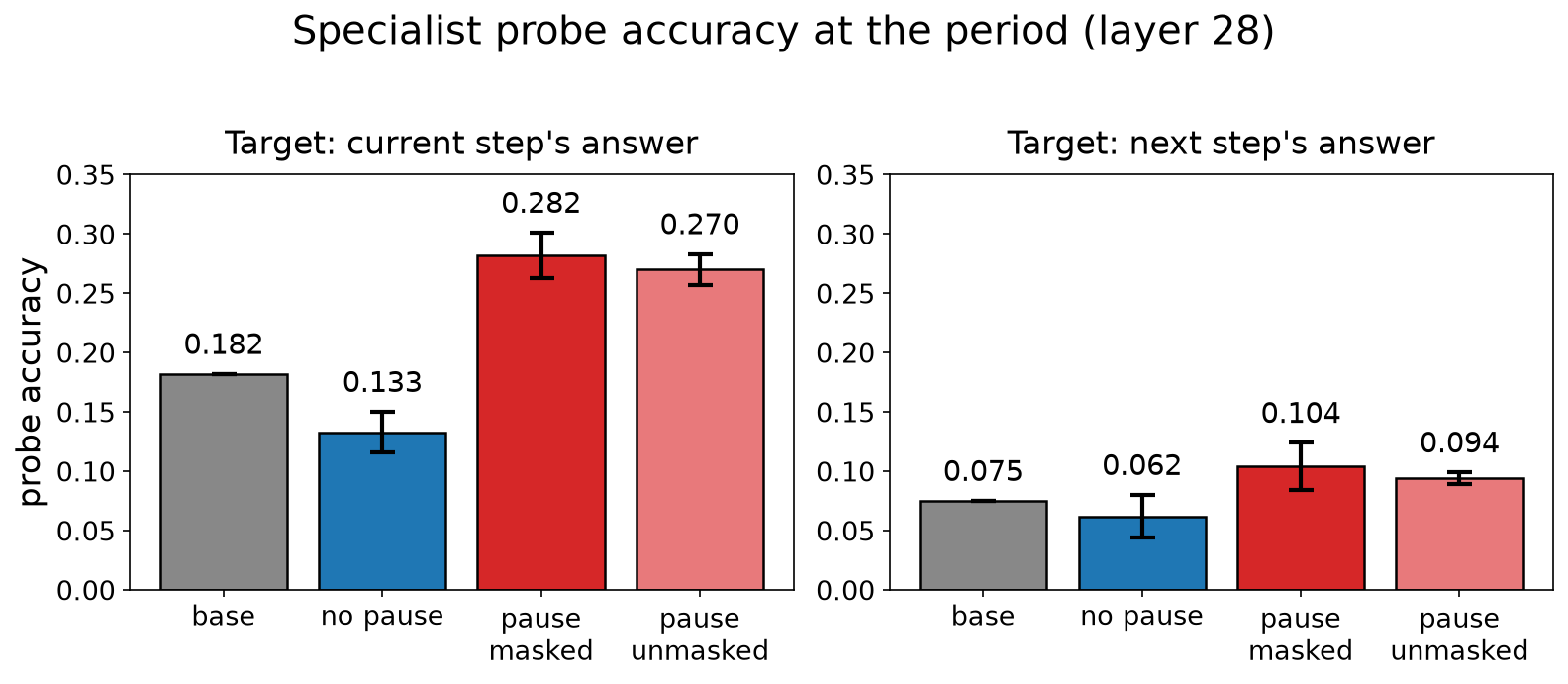}
    \vspace{-20pt}
    \caption{Linear-probe accuracy on the boundary-adjacent token's
    final hidden state ($n{=}3$ seeds, $\pm 1\sigma$).
    \textbf{Left:} current step's integer answer. \textbf{Right:}
    next step's integer answer. \textsc{Pause (unmasked)} trains the same boundary pauses as prediction targets.}
    \label{fig:probe_non_myopic}
\end{figure}

\paragraph{Results.}
Figure~\ref{fig:probe_non_myopic} shows masked-pause SFT yielding the highest decodability on both targets: current-step accuracy roughly doubles over no pause ($0.282$ vs.\ $0.133$, and clearly above the base $0.182$), and next-step accuracy is highest as well ($0.104$ vs.\ $0.062$ for no pause and $0.075$ for base). Training the same boundary pauses as prediction targets preserves most of the effect ($0.270$ and $0.094$), indicating that the compression follows primarily from pause placement, with masking a small additional increment. The pattern is consistent across seeds. 

\paragraph{Hypothesis.}
Under ordinary SFT, the period token must predict the first token of the next reasoning step; inserting a pause replaces that target with \texttt{[PAUSE]}, and masking additionally removes its loss. Either way the period's representation is freed from immediate next-token specialization and can instead absorb gradient from downstream positions:

\begin{quote}
\textbf{H2 (non-myopic compression).} \emph{Boundary pauses redirect gradient flow such that boundary-adjacent tokens encode more downstream information than under ordinary SFT.}
\end{quote}


\subsection{From hypotheses to predictions}
\label{sec:predictions}

Motivated by H1 and H2, the rest of the paper tests three predictions that follow if they hold at scale on real reasoning tasks:

\begin{itemize}
\setlength\itemsep{0pt}
\item \textbf{RQ1 (scaling).} Does masked boundary pause fine-tuning
at 1B--8B improve math and code reasoning while preserving general
capability (MMLU~\cite{hendrycks2020measuring}, GPQA~\cite{rein2024gpqa}, BBH~\cite{suzgun2022challenging}) better than standard fine-tuning?
H1 predicts this trade-off.

\item \textbf{RQ2 (placement).} Does pause placement affect performance at matched pause-token budget? If so, which strategy is optimal? H1 and H2 both act locally at the inserted positions, predicting that placement matters and that boundary insertion
outperforms random, prompt-append~\cite{goya}, and
low-confidence~\cite{learntoinsert} alternatives.



\item \textbf{RQ3 (RL exploration).} Do GRPO rollouts under MBP retain wider behavioral support than standard GRPO --- more per-prompt reward variance, slower distribution sharpening, and improved convergence? H1 predicts so.

\end{itemize}


\noindent Sections~\ref{sec:method}--\ref{sec:analysis} formalize
MBP and answer these questions via 1B--8B math/code SFT with OOD-retention measurement, and GRPO experiments.



\providecommand{\CC}{\mathcal{C}_1}
\providecommand{\SSet}{\mathcal{S}}
\providecommand{\SO}{\mathcal{S}_O}
\providecommand{\SP}{\mathcal{S}_P}
\providecommand{\VV}{\mathcal{V}}
\providecommand{\Wh}{\widehat{W}}
\providecommand{\Wo}{\widehat{W}_O}
\providecommand{\Wp}{\widehat{W}_P}
\providecommand{\ep}{e_p}
\providecommand{\eph}{\widehat{e}_P}
\providecommand{\Drift}{\Delta_1}
\providecommand{\Retain}{\mathrm{Retain}}
\providecommand{\Iop}{I_O}
\providecommand{\Ip}{I_1}
\providecommand{\Romax}{R_O}
\providecommand{\bctx}{\beta_{\mathrm{ctx}}}
\providecommand{\actx}{\alpha_{\mathrm{ctx}}}
\providecommand{\carry}{\mathrm{carry}}
\providecommand{\loc}{\mathrm{loc}}
\providecommand{\opnorm}[1]{\lVert #1 \rVert_{\infty\to\infty}}
\providecommand{\linfnorm}[1]{\lVert #1 \rVert_{\infty}}

\section{A retention certificate for masked-pause SFT}
\label{sec:theory}

H1 admits a first-order retention certificate. The intuition: old
$D_1$ predictions on no-pause prefixes survive $D_2$ fine-tuning
when their top-token margins outlast the logit movement induced by
adaptation. Bounding this movement uniformly over a protected
prefix set converts retention into a comparison between the
certified drifts of two procedures. Let $\Phi(r)$ denote the fraction
of protected $D_1$ prefixes whose original margin exceeds $r$.

\begin{theorem}[Retention certificate, informal]
\label{thm:pause_retention}
In the local linearized regime around the $D_1$-trained parameters
and matched to the same $D_2$ fitting level, let $\delta_O$ be the
certified protected drift of ordinary SFT on no-pause $D_1$ logits.
Masked-pause SFT has certified protected drift at most
$\beta_{\mathrm{ctx}}\delta_O$, where $\beta_{\mathrm{ctx}}$ measures
how separated the pause-perturbed prefix is from the ordinary prefix
in the Jacobian sense. Consequently,
\[
\begin{aligned}
\Retain(W_{\mathrm{SFT}})
&\ge \Phi(2\delta_O),\\
\Retain(W_{\mathrm{pause}})
&\ge \Phi(2\beta_{\mathrm{ctx}}\delta_O).
\end{aligned}
\]
Since $\Phi$ is non-increasing, whenever $\beta_{\mathrm{ctx}}<1$,
masked-pause SFT inherits a stronger retention guarantee than
ordinary SFT.
\end{theorem}

The condition $\beta_{\mathrm{ctx}} < 1$ has a geometric reading:
masked-pause SFT fits $D_2$ with less drift on $D_1$ no-pause
logits whenever the pause-perturbed state is sufficiently separated
from the ordinary state. This separation is structural rather than
architectural --- a property of \emph{where} pauses are inserted,
which the boundary selector of MBP encodes by design. The full
setup, lemma statements (margin stability, protected drift bounds),
and proofs are in Appendix~\ref{app:pause_retention},~\ref{app:pause_retention_proof}.





\section{Method}
\label{sec:method}

Motivated by H1 and H2 in \S\ref{sec:pilots}, we formalize the
masked pause training strategy that are predicted to exhibit optimal performance.

\subsection{Masked Boundary Pause Training}
\label{sec:mbp}
MBP specializes the position selector $g$ from \S\ref{sec:prelims}
to a \emph{boundary selector} $g_{\text{Boundary}}$ that produce segments $u_n$ of
Eq.~\ref{eq:segments} that are \emph{semantic units}: sentences,
reasoning steps, or code lines. Further details can be found in Appendix~\ref{app:step_boundary_extraction}.

Boundary placement is the advantageous choice for both hypotheses:
\begin{itemize}
\setlength\itemsep{2pt}
\item \textbf{H1 (mode retention).} Step boundaries are by construction
where semantic transitions occur, and we expect fine-tuning's
structural learning to concentrate there. A pause embedding at these positions absorbs that learning,
preserving the pretrained distribution 
that would otherwise drift (\S\ref{sec:pilot_mlc}).

\item \textbf{H2 (non-myopic compression).} Inserted pause positions
encourage implicit summarization at the preceding token~(\S\ref{sec:probing_pilot}), so
semantic boundaries are the natural placement: there the
compression encapsulates a complete semantic unit, rather than
being forced to predict the next token mid-sentence.
\end{itemize}

The MBP training loss is Eq.~\ref{eq:pause_loss} with
$g = g_{\text{Boundary}}$, and all placement baselines in
\S\ref{sec:sft_setup} use the same pause-token budget $M$ so that
comparisons isolate the role of $g$.

\subsection{Extension to GRPO}

MBP extends to GRPO without changing rollouts or rewards: responses
are sampled and scored on the clean policy, while pauses appear only
when computing the policy update. Given a sampled response $y_i$,
we form $\tilde y_i$ (Eq.~\ref{eq:augmented}) and apply the GRPO
objective at non-pause positions, with policy ratio and KL penalty
computed under the pause-perturbed prefix $(q, \tilde y_{i,<j})$
rather than the clean prefix $(q, y_{i,<j})$. Pauses thus remain
outside the action space, perturbing only the policy update's
context --- the mechanism through which we test RQ3
(\S\ref{sec:predictions}): whether retention of the pretrained
distribution improves GRPO's exploration.

\section{Experiment Setup}
\label{sec:experiment_setup}

We test the predictions of \S\ref{sec:predictions} in two settings:
downstream math/code supervised fine-tuning (\S\ref{sec:sft_setup},
RQ1 and RQ2) and GRPO-based reasoning training
(\S\ref{sec:grpo_setup}, RQ3). Full hyperparameters and data details
are in Appendix~\ref{app:experimental_settings}.








\subsection{Downstream Math/Code SFT}
\label{sec:sft_setup}

\noindent\textbf{Setup.} We fine-tune Qwen3 (1.7B, 4B, 8B-Base)\cite{yang2025qwen3} and Llama-3.1-8B-Instruct~\cite{grattafiori2024llama} on
math data (83K curated DeepMath-103K~\cite{deepmath103k} subset) and code data (100K
OpenCodeInstruct subset~\cite{ahmad2025opencodeinstruct}), inserting
pauses only on the response side. We apply LoRA~\cite{hu2022lora} for 8B models. We evaluate reasoning on math
(GSM8K, MATH, AMC23, AIME24, Minerva)~\cite{cobbe2021training, hendrycks2021measuring,lewkowycz2022minerva} and code (HumanEval(+),
MBPP(+))~\cite{chen2021humaneval,austin2021mbpp,liu2023evalplus} benchmarks; general-capability retention on MMLU, GPQA,
HellaSwag, BBH, PIQA~\cite{hendrycks2020measuring,rein2024gpqa,zellers2019hellaswag,suzgun2022challenging,bisk2020piqa}. We report pass@1 for math and code, and pass@4
for general-capability benchmarks.

\noindent\textbf{Baselines (matched pause-token budget).}
\textsc{SFT} (no pauses); \textsc{Append}: pauses at the
prompt-response boundary~\cite{goya}; \textsc{Random}: pauses at
random response positions; \textsc{DIT}: pauses before
low-likelihood target tokens~\cite{learntoinsert}; \textsc{MBP}
(ours): pauses at reasoning-step or code-line boundaries. All pause
variants are loss-masked except \textsc{MBP w/o masking}, which
supervises pause tokens as targets (\S\ref{sec:retention_analyses}).
Matching the budget isolates placement from token count.

\subsection{GRPO Setup}
\label{sec:grpo_setup}

We compare \textsc{Standard GRPO} against \textsc{MBP GRPO} on Qwen3-1.7B and 4B, training on 5K-problem subsets of DeepMath-103K~\cite{deepmath103k} and SimpleRLZoo~\cite{simplerlzoo}. We report pass@1 on the math benchmarks above, and track two training-dynamics metrics: convergence AUC (area under the training reward curve, measuring how quickly reward accumulates), and mixed-group ratio (proportion of rollout groups containing both correct and incorrect rollouts).

\begin{table}[t]
    \centering
    \begingroup
    \scriptsize
    \setlength{\tabcolsep}{2.1pt}
    \renewcommand{\arraystretch}{1.05}
    \resizebox{\columnwidth}{!}{%
    \begin{tabular}{llcccccc}
    \toprule
    \textbf{Model}
    & \textbf{Method}
    & \textbf{GSM8K}
    & \textbf{MATH500}
    & \textbf{MATH}
    & \textbf{AMC23}
    & \textbf{MINERVA}
    & \textbf{Avg.} \\
    \midrule
    
    \multirow{3}{*}{\shortstack[c]{Qwen3-1.7B\\Base}}
    & Base           & 65.13 & 54.20 & 54.82 & 35.00 & 17.65 & 45.36 \\
    & SFT            & 82.41 & \textbf{63.40} & 62.98 & 27.50 & 20.59 & 51.38 \\
    \cmidrule(lr){2-8}
    & \textbf{MBP}   & \textbf{83.02} & 61.00 & \textbf{64.56} & \textbf{37.50} & \textbf{23.16} & \textbf{53.85} \\
    
    \midrule
    
    \multirow{3}{*}{\shortstack[c]{Qwen3-4B\\Base}}
    & Base           & 82.56 & 66.00 & 67.64 & 37.50 & 28.68 & 56.48 \\
    & SFT            & \textbf{91.36} & 69.20 & 71.24 & 45.00 & 29.78 & 61.32 \\
    \cmidrule(lr){2-8}
    & \textbf{MBP}   & 90.07 & \textbf{73.40} & \textbf{72.32} & \textbf{50.00} & \textbf{30.51} & \textbf{63.26} \\
    
    \midrule
    
    \multirow{3}{*}{\shortstack[c]{Qwen3-8B\\Base}}
    & Base           & 88.40 & 65.40 & 65.86 & 42.50 & 26.84 & 57.80 \\
    & SFT            & \textbf{92.95} & 71.60 & 72.54 & 52.50 & 32.72 & 64.46 \\
    \cmidrule(lr){2-8}
    & \textbf{MBP}   & 92.57 & \textbf{75.00} & \textbf{72.84} & \textbf{55.00} & \textbf{35.29} & \textbf{66.14} \\
    
    \midrule
    
    \multirow{3}{*}{\shortstack[c]{Llama3.2-1B\\Instruct}}
    & Base           & 38.97 & 23.80 & 25.70 & 15.00 & 7.72 & 22.24 \\
    & SFT            & 50.72 & 28.00 & 38.48 & 10.00 & 8.09 & 27.06 \\
    \cmidrule(lr){2-8}
    & \textbf{MBP}   & \textbf{51.63} & \textbf{31.80} & \textbf{41.66} & \textbf{27.50} & \textbf{14.34} & \textbf{33.39} \\
    
    \midrule
    
    \multirow{3}{*}{\shortstack[c]{Llama3.2-3B\\Instruct}}
    & Base           & 70.43 & 40.00 & 43.22 & 10.00 & 17.65 & 36.26 \\
    & SFT            & 76.80 & 46.40 & 55.02 & 27.50 & 13.97 & 43.94 \\
    \cmidrule(lr){2-8}
    & \textbf{MBP}   & \textbf{77.03} & \textbf{48.20} & \textbf{55.92} & \textbf{30.00} & \textbf{19.85} & \textbf{46.20} \\
    
    \midrule
    
    \multirow{3}{*}{\shortstack[c]{Llama3.1-8B\\Instruct}}
    & Base           & 77.48 & 46.40 & 48.62 & \underline{26.50} & \textbf{22.43} & 44.29 \\
    & SFT            & \underline{84.84} & \underline{53.20} & \underline{56.28} & 17.50 & \underline{21.69} & \underline{46.70} \\
    \cmidrule(lr){2-8}
    & \textbf{MBP}   & \textbf{86.28} & \textbf{58.00} & \textbf{61.02} & \textbf{27.50} & \textbf{22.43} & \textbf{51.05} \\
    \bottomrule
    
    \end{tabular}%
    }
    \endgroup
    \caption{
    Math-domain results.
    Average denotes the macro-average over GSM8K, MATH500, MATH, AMC23, and MINERVA.
    }
    \label{tab:math_domain_results}
    \end{table}

\begin{table}[t]
    \centering
    \begingroup
    \scriptsize
    \setlength{\tabcolsep}{2.6pt}
    \renewcommand{\arraystretch}{1.05}
    \resizebox{\columnwidth}{!}{%
    \begin{tabular}{llccccc}
    \toprule
    \textbf{Model}
    & \textbf{Method}
    & \textbf{MBPP}
    & \textbf{MBPP+}
    & \textbf{HumanEval}
    & \textbf{HumanEval+}
    & \textbf{Avg.} \\
    \midrule
    
    \multirow{3}{*}{\shortstack[c]{Qwen3-1.7B\\Base}}
    & Base           & 69.05 & 59.26 & 58.54 & 53.05 & 59.98 \\
    & SFT            & 69.58 & \textbf{62.43} & 67.07 & 61.59 & 65.17 \\
    \cmidrule(lr){2-7}
    & \textbf{MBP}   & \textbf{73.02} & \textbf{62.43} & \textbf{70.12} & \textbf{65.24} & \textbf{67.70} \\
    
    \midrule
    
    \multirow{3}{*}{\shortstack[c]{Qwen3-4B\\Base}}
    & Base           & 74.34 & 64.81 & 76.83 & 72.56 & 72.14 \\
    & SFT            & 78.04 & \textbf{67.99} & 79.27 & 73.78 & 74.77 \\
    \cmidrule(lr){2-7}
    & \textbf{MBP}   & \textbf{79.10} & 65.08 & \textbf{82.93} & \textbf{78.05} & \textbf{76.29} \\
    
    \midrule
    
    \multirow{3}{*}{\shortstack[c]{Qwen3-8B\\Base}}
    & Base           & 82.80 & 72.49 & \textbf{88.41} & \textbf{84.15} & 81.96 \\
    & SFT            & 86.51 & 75.13 & 85.98 & 81.71 & 82.33 \\
    \cmidrule(lr){2-7}
    & \textbf{MBP}   & \textbf{87.30} & \textbf{75.66} & 86.59 & 82.32 & \textbf{82.97} \\
    
    \midrule
    
    \multirow{3}{*}{\shortstack[c]{Llama3.1-8B\\Instruct}}
    & Base           & 67.46 & 59.52 & 67.68 & 62.20 & 64.22 \\
    & SFT            & 68.25 & \textbf{60.05} & 68.29 & \textbf{64.02} & 65.15 \\
    \cmidrule(lr){2-7}
    & \textbf{MBP}   & \textbf{68.52} & \textbf{60.05} & \textbf{69.51} & \textbf{64.02} & \textbf{65.53} \\
    \bottomrule

    \end{tabular}%
    }
    \endgroup
    \caption{
    Code-domain results.
    Average denotes the macro-average over MBPP, MBPP+, HumanEval, and HumanEval+.
    }
    \label{tab:code_domain_results}
\end{table}

\section{Results}
\label{sec:results}

We answer RQ1 (\S\ref{sec:scaling_results}), RQ2 (\S\ref{sec:placement_results}), and RQ3 (\S\ref{sec:grpo_results}) in order. Across these results, the findings show the patterns predicted by H1 and H2.

\subsection{Scaling and Retention (RQ1)}
\label{sec:scaling_results}
\paragraph{Reasoning gains scale across model sizes and families.}
Tables~\ref{tab:math_domain_results} and~\ref{tab:code_domain_results}
show that MBP outperforms standard SFT on every model variant we
tested. Math gains over SFT range from $+1.7$ to $+6.3$ points
across Qwen3 (1.7B-8B-Base) and Llama (3.2-1B/3B-Instruct,
3.1-8B-Instruct); code gains range from $+0.6$ to $+2.5$ across
Qwen3. The improvement is consistent in sign across every model
variant, showing that pause training is not a model-specific phenomenon.

\paragraph{General capability is preserved.}
Table~\ref{tab:general_domain_results} reports MMLU, GPQA, HellaSwag,
BBH, and PIQA on Qwen3-1.7B-Base after math and code SFT. Standard
SFT is at best neutral on general capability after math SFT and degrades it after code SFT. By contrast, MBP improves the general-benchmark average over base in both settings ($+4.2$ after math, $+2.5$ after code), beating SFT by $+3.1$ and $+4.8$, respectively. The MBP-vs-SFT gap on general benchmarks is larger than the gap on the trained-domain benchmarks themselves. This suggests that MBP does not trade reasoning gains for retention loss, but improves both simultaneously.

These findings support H1 at scale: MBP preserves the pretrained
distribution while still acquiring new-task reasoning.

\begin{table}[t]
\centering
\begingroup
\scriptsize
\setlength{\tabcolsep}{2.2pt}
\renewcommand{\arraystretch}{1.05}
\resizebox{\columnwidth}{!}{%
\begin{tabular}{llcccccc}
\toprule
\textbf{Trained Domain}
& \textbf{Method}
& \textbf{MMLU}
& \textbf{GPQA}
& \textbf{HellaSwag}
& \textbf{BBH}
& \textbf{PIQA}
& \textbf{Avg.} \\
\midrule

\multirow{3}{*}{Math}
& Base           & 62.95 & \textbf{27.46} & 38.69 & 38.84 & 62.13 & 46.01 \\
& SFT            & 62.32 & 24.78 & 40.39 & 42.06 & 66.10 & 47.13 \\
\cmidrule(lr){2-8}
& \textbf{MBP}   & \textbf{63.93} & 27.23 & \textbf{48.32} & \textbf{44.13} & \textbf{67.63} & \textbf{50.25} \\

\midrule

\multirow{3}{*}{Code}
& Base           & 62.95 & \textbf{29.91} & 38.00 & 38.84 & 62.13 & 46.37 \\
& SFT            & 60.19 & 23.88 & 36.44 & 39.15 & 60.66 & 44.06 \\
\cmidrule(lr){2-8}
& \textbf{MBP}   & \textbf{63.78} & 25.67 & \textbf{44.01} & \textbf{42.88} & \textbf{68.01} & \textbf{48.87} \\
\bottomrule
\end{tabular}%
}
\endgroup
\caption{
General-domain results on Qwen3-1.7B-Base.
Average denotes the macro-average over MMLU, GPQA, HellaSwag, BBH, and PIQA.
}
\label{tab:general_domain_results}
\end{table}

\subsection{Placement and Masking (RQ2)}
\label{sec:placement_results}

H1 and H2 predict that boundary placement activates both
mechanisms: placements without alignment to semantic transitions
(append, random) or correlated with hard tokens rather than
transitions (DIT) should fail to do so. We test this on
Qwen3-1.7B-Base at matched pause-token budget.

\paragraph{Boundary placement is the only consistent winner.}
Table~\ref{tab:method_ablation_results} reveals a $7.5$-point spread
across the four placements on math (from DIT's $-5.0$ to MBP's
$+2.5$ relative to SFT), despite identical pause-token budgets. Of
the four, only MBP is reliably positive on both math and code domains. The other placements fluctuate: Append and DIT modestly help code ($+1.4$, $+1.7$) but hurt math ($-0.9$,
$-5.0$). Random is near-flat on code ($+0.2$) and clearly negative
on math ($-1.7$). The inconsistency suggests that random, append,
and low-confidence positions activate H1 and H2's mechanisms only
when an inserted pause happens to coincide with a meaningful
boundary, whereas \textsc{Boundary} placement targets these
positions by construction and delivers reliable gains across
domains.

    \begin{table}[t]
    \centering
    \begingroup
    \scriptsize
    \setlength{\tabcolsep}{2.7pt}
    \renewcommand{\arraystretch}{1.05}
    
    \resizebox{\columnwidth}{!}{%
    \begin{tabular}{lcccccc}
    \toprule
    \multirow{2}{*}{\textbf{Method}}
    & \multicolumn{5}{c}{\textbf{Math}}
    & \multirow{2}{*}{\textbf{Avg.}} \\
    \cmidrule(lr){2-6}
    & \textbf{GSM8K}
    & \textbf{MATH500}
    & \textbf{MATH}
    & \textbf{AMC23}
    & \textbf{MINERVA}
    & \\
    \midrule
    Base               & 65.13 & 54.20 & 54.82 & \underline{35.00} & 17.65 & 45.36 \\
    SFT                & \underline{82.41} & \textbf{63.40} & 62.98 & 27.50 & 20.59 & \underline{51.38} \\
    Append             & 79.68 & 59.20 & 60.96 & 30.00 & \underline{22.79} & 50.53 \\
    Random             & 81.35 & 60.80 & \underline{63.26} & 20.00 & \textbf{23.16} & 49.71 \\
    DIT                & 68.01 & 57.20 & 56.38 & 30.00 & 20.22 & 46.36 \\
    \cmidrule(lr){1-7}
    \textbf{MBP (ours)}      & \textbf{83.02} & 61.00 & \textbf{64.56} & \textbf{37.50} & \textbf{23.16} & \textbf{53.85} \\
    \textbf{MBP w/o masking} & 80.97 & \underline{62.20} & 62.64 & 27.50 & \textbf{23.16} & 51.29 \\
    \bottomrule
    \end{tabular}%
    }
    
    \vspace{0.35em}
    
    \resizebox{\columnwidth}{!}{%
    \begin{tabular}{lccccc}
    \toprule
    \multirow{2}{*}{\textbf{Method}}
    & \multicolumn{4}{c}{\textbf{Code}}
    & \multirow{2}{*}{\textbf{Avg.}} \\
    \cmidrule(lr){2-5}
    & \textbf{MBPP}
    & \textbf{MBPP+}
    & \textbf{HumanEval}
    & \textbf{HumanEval+}
    & \\
    \midrule
    Base               & 69.05 & 59.26 & 58.54 & 53.05 & 59.98 \\
    SFT                & 69.58 & \textbf{62.43} & 67.07 & 61.59 & 65.17 \\
    Append             & 69.84 & 60.58 & \underline{70.73} & \textbf{65.24} & 66.60 \\
    Random             & \underline{70.37} & \underline{61.11} & 67.07 & \underline{62.80} & 65.34 \\
    DIT                & 69.84 & 60.58 & \textbf{71.95} & \textbf{65.24} & \underline{66.90} \\
    \cmidrule(lr){1-6}
    \textbf{MBP (ours)}      & \textbf{73.02} & \textbf{62.43} & 70.12 & \textbf{65.24} & \textbf{67.70} \\
    \textbf{MBP w/o masking} & 70.63 & 58.99 & 67.68 & 62.20 & 64.88 \\
    \bottomrule
    \end{tabular}%
    }
    
    \endgroup
    \caption{
    Method ablation results on Qwen3-1.7B-Base.
    The upper block reports math benchmarks, and the lower block reports code benchmarks.
    }
    \label{tab:method_ablation_results}
    \end{table}
    
\subsection{Extension to GRPO exploration (RQ3)}
\label{sec:grpo_results}

If MBP preserves the pretrained distribution (H1), GRPO rollouts
should retain wider behavioral support, yielding a richer gradient
signal. We test this on Qwen3-1.7B-Base and Qwen3-4B-Base.


\paragraph{Faster convergence, higher final accuracy.}
Table~\ref{tab:grpo_accuracy_auc} shows MBP-GRPO improves final
four-benchmark average accuracy by $+1.31$ points (Qwen3-1.7B) and
$+0.40$ points (Qwen3-4B) over standard GRPO, while normalized
convergence AUC increases by $+0.048$ and $+0.045$, respectively.

\begin{figure}[t]
    \centering
    \includegraphics[width=1.0\linewidth]{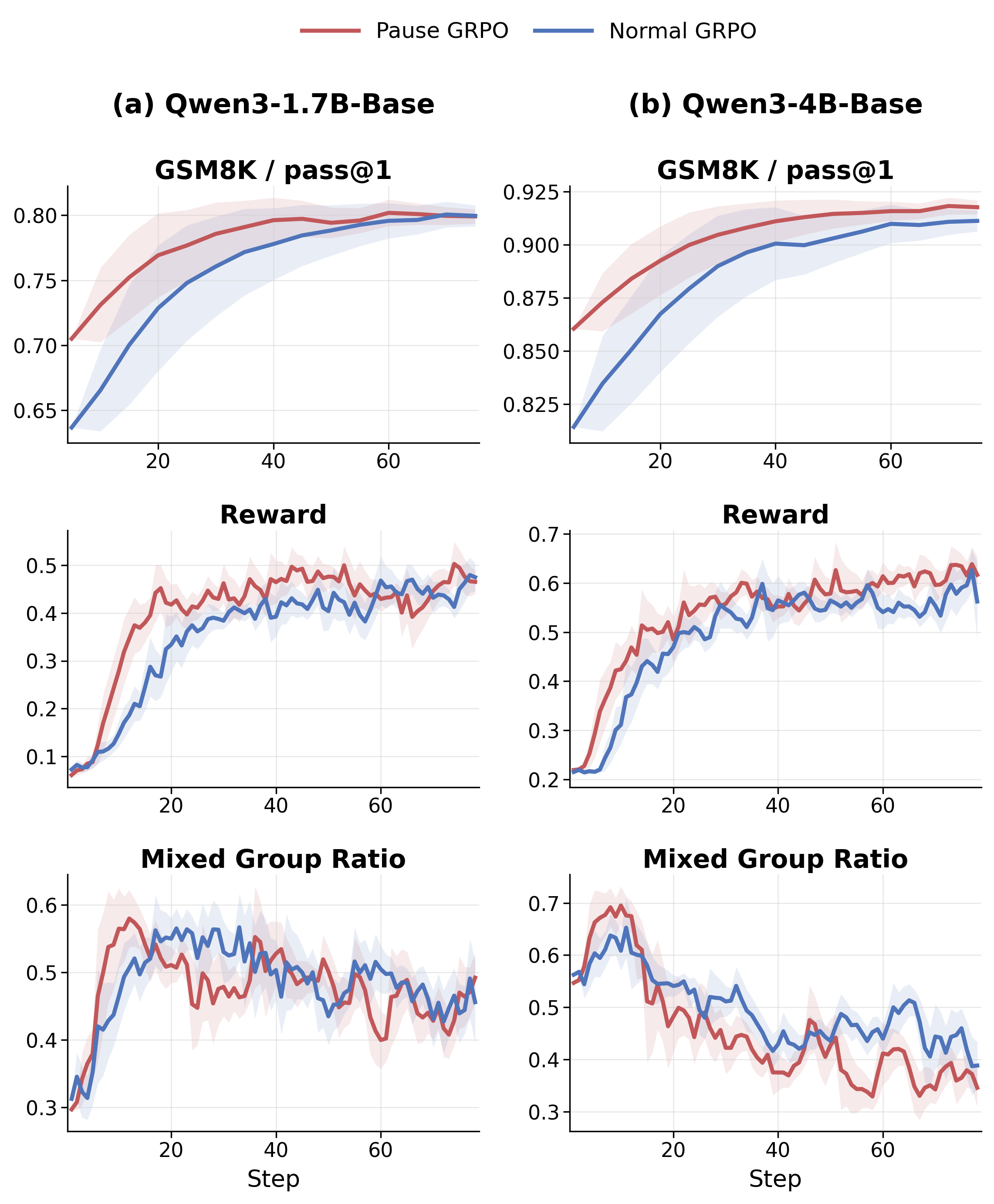}
    \vspace{-20pt}
    \caption{\textbf{Training dynamics of MBP-GRPO.}
        Columns correspond to Qwen3-1.7B-Base (left) and Qwen3-4B-Base (right), while rows show GSM8K validation pass@1 (top), training reward (middle), and mixed-group ratio (bottom). Red denotes MBP-GRPO and blue denotes standard GRPO. MBP-GRPO exhibits a higher mixed-group ratio in the early stage of training, providing more useful update signals and leading to faster convergence than standard GRPO.}
    \label{fig:grpo_stacked_graph}
\end{figure}

\paragraph{Early rollout diversity drives the gain.}
Figure~\ref{fig:grpo_stacked_graph} shows that during the
earliest steps, MBP-GRPO's mixed-group ratio spikes substantially
above standard GRPO, generating more useful gradient updates when
the advantage signal is most valuable. This can be observed in reward and accuracy curves. MBP-GRPO pulls ahead from the
first steps and stays ahead throughout training. Later, MBP's
mixed-group ratio falls below standard's, consistent with earlier
convergence on solved prompts. The pattern matches H1 in that a retained
pretrained distribution provides more diverse rollouts when they
most help, and the early advantage compounds.

\begin{table}[t]
\centering
\small
\setlength{\tabcolsep}{2.5pt}
\renewcommand{\arraystretch}{1.0}
\begin{tabular}{@{}lccccc@{}}
\toprule
\textbf{Method} & \textbf{GSM8K} & \textbf{MATH} & \textbf{AMC23} & \textbf{Minerva} & \textbf{AUC} \\
\modelrow{6}{Qwen3-1.7B-Base}
\textsc{Standard} & 80.52 & 59.68 & 37.50 & 20.96 & 0.366 \\
\oursrule{6}
\textbf{\textsc{MBP}} & \textbf{81.35} & \textbf{61.24} & \textbf{40.00} & \textbf{21.32} & \textbf{0.414} \\
\modelrow{6}{Qwen3-4B-Base}
\textsc{Standard} & \textbf{92.19} & 74.30 & 50.00 & \textbf{30.15} & 0.505 \\
\oursrule{6}
\textbf{\textsc{MBP}} & 91.58 & \textbf{75.26} & \textbf{57.50} & 23.90 & \textbf{0.550} \\
\bottomrule
\end{tabular}
\caption{Final GRPO pass@1 accuracy (\%) on GSM8K, MATH, AMC23, and Minerva, together with normalized training-reward AUC, for Qwen3-1.7B-Base and Qwen3-4B-Base. Full results are reported in Appendix Table~\ref{tab:grpo_accuracy_auc_full}.}
\label{tab:grpo_accuracy_auc}
\end{table}

\section{Analysis}
\label{sec:analysis}

\subsection{Retention Analyses}
\label{sec:retention_analyses}

We examine H1's retention mechanism through a decomposition of
post-training correctness, an ablation on loss-masking, and
next-token entropy at step boundaries.

\paragraph{MBP preserves base-correct examples while solving new ones.}
If H1 holds at scale, MBP should preserve pretrained capabilities while fitting to the downstream distribution. We measure
both directly: BCP (Base-Correct Preservation) is the fraction of
base-correct examples still correct after fine-tuning, and NSR
(New-Solve Rate) is the fraction of base-wrong examples that become
correct. Table~\ref{tab:bcp_nsr_acc_standard_mbp} shows MBP improves both BCP and NSR.
This shows that MBP achieves sufficient adaptation while better preserving the pretrained distribution.



\begin{table}[t]
\centering
\small
\setlength{\tabcolsep}{6pt}
\renewcommand{\arraystretch}{1.05}
\begin{tabular}{@{}llccc@{}}
\toprule
Metric & Method & GSM8K & MATH & Overall \\
\midrule
\multirow{2}{*}{BCP $\uparrow$}
& \textsc{Standard} & 92.08 & 88.33 & 89.18 \\
& \textsc{MBP}      & \textbf{93.36} & \textbf{89.09} & \textbf{90.00} \\
\midrule
\multirow{2}{*}{NSR $\uparrow$}
& \textsc{Standard} & 63.04 & 33.42 & 38.20 \\
& \textsc{MBP}      & \textbf{66.74} & \textbf{36.83} & \textbf{41.11} \\
\midrule
\multirow{2}{*}{Acc. $\uparrow$}
& \textsc{Standard} & 81.96 & 63.52 & 67.14 \\
& \textsc{MBP}      & \textbf{84.08} & \textbf{65.48} & \textbf{68.87} \\
\bottomrule
\end{tabular}
\caption{BCP(Base-Correct Preservation), NSR(New-Solve Rate), and final accuracy of Standard-SFT and MBP on Qwen3-1.7B-Base.}
\label{tab:bcp_nsr_acc_standard_mbp}
\end{table}

\paragraph{Masking is essential at boundary placement.}
We predict that masking pause tokens would be advantageous in retaining the pretrained distribution. To test this, we perform ablation experiments in math and code training where pauses are predicted as targets~\cite{thinksentence}. Table~\ref{tab:method_ablation_results} (\textsc{MBP w/o masking}) shows that switching from masked to
predicted costs $-2.6$ points on math and $-2.8$ on code; on code, predicted boundary placement even falls
below standard SFT, confirming that masking is load-bearing.

\paragraph{Entropy is retained at step boundaries.}

H1 predicts that pause training selectively preserves diversity at boundaries.
We therefore measure entropy both at step-boundary positions and across all token positions on out-of-distribution (OOD) general language understanding benchmarks, with the in-distribution GSM8K and MATH benchmarks included for comparison (Table~\ref{tab:boundary_entropy}).
At step boundaries, MBP maintains substantially higher entropy than SFT across both in-domain and OOD tasks, suggesting that MBP selectively preserves the base model’s distributional diversity at these positions.
In contrast, the average entropy over all tokens remains almost identical between SFT and MBP.
Taken together, these results indicate that MBP selectively preserves distributional diversity at step boundaries while committing to a distribution within steps.


\begin{table}[t]
\centering
\small
\setlength{\tabcolsep}{3.2pt}
\renewcommand{\arraystretch}{1.0}
\begin{tabular}{@{}lcccc@{}}
\toprule
& \multicolumn{2}{c}{\textbf{OOD}} & \multicolumn{2}{c}{\textbf{In-domain}} \\
\cmidrule(lr){2-3}\cmidrule(lr){4-5}
\textbf{Model} & \textbf{MMLU} & \textbf{GPQA} & \textbf{GSM8K} & \textbf{MATH} \\
\midrule
\multicolumn{5}{@{}l}{\emph{Step boundaries}} \\
Base & 0.707 & 0.607 & 0.500 & 0.579 \\
SFT  & 0.403 & 0.380 & 0.278 & 0.339 \\
MBP  & 0.463 & 0.440 & 0.365 & 0.380 \\
\midrule
\multicolumn{5}{@{}l}{\emph{All tokens}} \\
Base & 0.587 & 0.489 & 0.358 & 0.258 \\
SFT  & 0.361 & 0.321 & 0.189 & 0.144 \\
MBP  & 0.368 & 0.332 & 0.192 & 0.145 \\
\bottomrule
\end{tabular}
\caption{Next-token vocabulary entropy on Qwen3-1.7B-Base rollouts under the Base, math-SFT, and MBP checkpoints, measured at step boundaries and across all response tokens. MMLU and GPQA are out of domain; GSM8K and MATH are in domain.}
\label{tab:boundary_entropy}
\end{table}

\subsection{Non-Myopic Compression}
\label{sec:non_myopic_compression}

An intuitive method to empirically quantify the extent to which H2 holds would be through mutual information (MI) measurements between step-boundary tokens and the final answer. We propose an MI proxy,
\begin{equation*}
\begin{split}
I(X;Y) = \mathbb{E}_{(x,y)\sim p_\theta(\cdot,\cdot\mid q)}
\biggl[ \log \frac{p_\theta(x,y\mid q)}{p_\theta(x\mid q)\,p_\theta(y\mid q)} \biggr]
\end{split}
\end{equation*}

where x is the intermediate reasoning steps, y is the final answer, and q is the input prompt. 

This measure evaluates the predictability of the final answer from the intermediate step boundary, which reflects the extent to which boundary tokens compress final answer-related information (Appendix~\ref{app:mi_proxy_computation}). Table~\ref{tab:mi_proxy} compares the MI proxy values for MBP and standard SFT across mathematical reasoning benchmarks. Results show that MBP compresses significantly more final answer related information compared to boundaries from standard SFT, an empirical observation that is consistent with H2.


\begin{table}[t]
\centering
\small
\setlength{\tabcolsep}{6pt}
\renewcommand{\arraystretch}{1.0}
\begin{tabular}{@{}lccc@{}}
\toprule
\textbf{Dataset} & \textbf{Base} & \textbf{SFT} & \textbf{MBP} \\
\midrule
GSM8K        & 0.00 & 1.43 & \textbf{6.31} \\
MATH         & 0.00 & 2.64 & \textbf{4.84} \\
DeepMath Eval & 0.00 & 4.67 & \textbf{8.29} \\
\bottomrule
\end{tabular}
\caption{Base-referenced normalized MI proxy between context-conditioned step-boundary tokens and final answers. Results use macro aggregation over the full evaluation sets.}
\label{tab:mi_proxy}
\end{table}

\section{Related Work}
\paragraph{Pause and Dummy Tokens.} Prior works have utilized pause tokens to enhance LLM reasoning by delaying generation~\cite{goya,thinktokens}, and \citet{pausetoken} prove that pause tokens strictly increase the expressivity of constant-depth transformers. Building on this, recent approaches have explored dynamic token insertion at strategic points, like low-confidence positions or sentence boundaries, to better guide cognitive processing~\cite{learntoinsert,thinksentence}. Unlike these prior works that primarily view token insertion as a mechanism for increasing computational bandwidth or providing direct generation cues, our work uniquely isolates the role of pause tokens as training-time context perturbations.

\paragraph{Post-Training for Reasoning and Alignment.} Contemporary post-training paradigms for LLMs aim to align behavior and enhance complex problem-solving, typically progressing from SFT to reinforcement learning frameworks such as GRPO~\cite{zeroshot,instructionft,deepseekmath}. A critical challenge across these stages is the trade-off between reasoning adaptation and mode retention, as fine-tuning on a new reasoning task often degrades the model's previously established language fluency and factual knowledge~\cite{empiricalcf,spuriousforgetting,taxrlhf}. Our work is complementary to these post-training paradigms. We show that a lightweight token-insertion strategy can improve both SFT and GRPO when applied at semantic boundaries and masked from direct prediction, while mitigating the adaptation-retention trade-off.

\section{Conclusions}

In this work, we extend the study of pause tokens to their effect on fine-tuning dynamics. Through controlled pilot experiments, we identify two hypotheses, mode retention (H1) and non-myopic compression (H2), which naturally lead to Masked Boundary Pause Training. Through this lens, we show improvements in math and code reasoning while preserving general capability on unrelated benchmarks. We further show that boundary placement reliably outperforms SFT across model scales, as our hypotheses predict, and that retention also helps exploration in RL training. We hope this training-dynamics perspective, a shift from computational expressivity to train-time mode retention and non-myopic compression, supports further research across domains and adaptation regimes.

\section*{Limitations}
Our current study focuses on full fine-tuning and evaluates retention primarily through per-example correctness, using BCP and NSR to measure whether base-correct examples remain correct and base-wrong examples become solved. Additionally, initial LoRA experiments show more moderate gains, suggesting that MBP's interaction with parameter-efficient fine-tuning requires further study. Future work should examine retention at finer token- or distribution-level granularity and systematically analyze MBP under parameter-efficient adaptation.





\section*{Acknowledgments}
This work is in part supported by the National Research Foundation of Korea (NRF, RS-2024-00451435(15\%), RS-2024-00413957(15\%)), Institute of Information \& Communications Technology Planning \& Evaluation (IITP, RS-2025-02305453(10\%), RS-2025-02273157(10\%), RS-2025-25442149(20\%), RS-2021-II211343(10\%)) grant funded by the Ministry of Science and ICT (MSIT), Institute of New Media and Communications (INMAC), the BK21 FOUR program of the Education, Artificial Intelligence Graduate School Program (Seoul National University), Research Program for Future ICT Pioneers, Seoul National University in 2026, and the KT (Korea Telecom)--(Seoul National University/Korea University) AICT R\&D Center (20\%).

\bibliography{custom}

\clearpage


\appendix

\providecommand{\pausetok}{\textbf{\texttt{\char60{}|PAUSE|\char62{}}}}
\section{Pilot Experiment Details}
\label{app:pilot}

\subsection{MLC continual learning pilot}
\label{app:pilot_mlc}

\paragraph{Task and data.}
Sequences are generated by the Modular Linear Chain (MLC) recurrence
$x_{t+1} = (a_t\, x_t + b_t) \bmod p$ with prime $p = 37$ and
sequence length $L = 8$. A \emph{regime} is a fixed length-$L$
coefficient sequence $\{(a_t, b_t)\}_{t=1}^L$ with each $a_t, b_t$
drawn uniformly from $\{0, \ldots, p-1\}$. We sample two disjoint
coefficient sets to define regimes $D_1$ and $D_2$. Vocabulary size
is $p + 2$ (digits, BOS, and one pause token).

\paragraph{Model and optimizer.}
A 2-layer, 4-head GPT-style Transformer with hidden size $d = 128$
and learned positional embeddings. Trained with AdamW (learning
rate $3 \times 10^{-4}$, weight decay $0.01$), batch size $64$.

\paragraph{Two-phase training.}
\emph{Phase 1}: train on $D_1$ with standard next-token prediction
until $D_1$ accuracy exceeds $0.99$ (approximately $2{,}000$
optimizer steps). \emph{Phase 2}: fork from a single Phase-1
checkpoint and continue training on $D_2$ under two matched
conditions (\textsc{No pause} and \textsc{Pause}), each run for an
additional $2{,}000$ steps. The \textsc{Pause} condition inserts
one pause token after every content token, doubling sequence
length; pause-target positions are excluded from the cross-entropy
loss. Both conditions share architecture, optimizer, batch size,
learning rate, and update budget. We report $n = 8$ seeds with
shaded bands at $\pm 1\sigma$.

\paragraph{Evaluation.}
We track held-out loss on $D_2$ (adaptation) and $D_1$ (retention)
every $5$ Phase-2 steps. $D_2$ loss is computed on pause-augmented
sequences for \textsc{Pause} and bare sequences for \textsc{No
pause}; $D_1$ loss is bare for both conditions (matching the
no-pause inference distribution).

\subsection{Multi-stage continual learning}
\label{app:pilot_mlc_multistage}

The main pilot (\S\ref{sec:pilot_mlc}) uses a single $D_1 \to D_2$
transition. To test whether the retention advantage is specific to one
adaptation or persists as adaptations accumulate, we extend the same
setup to a four-regime chain.

\paragraph{Setup.}
We sample four disjoint coefficient sets defining regimes $D_1, \ldots,
D_4$ ($p = 37$, $L = 8$); all task, model, and optimizer settings follow
Appendix~\ref{app:pilot_mlc}. After Phase-1 training on $D_1$, each
condition adapts sequentially to $D_2$, $D_3$, and $D_4$ for $2{,}000$
steps per stage under matched budget and optimizer settings:
\textsc{No pause} runs standard SFT on each bare $D_k$, and
\textsc{Pause} runs masked-pause SFT. We report $n = 8$ seeds and track
held-out bare-format $D_1$ loss.

\paragraph{Results.}
Table~\ref{tab:pilot_mlc_multistage} reports held-out $D_1$ loss at the
end of each stage. The retention gap is positive in $8/8$ seeds at every
stage and never inverts across the $6{,}000$ steps, holding at roughly
$3$--$4$ nats throughout the chain. The absolute advantage therefore
neither accumulates nor decays as adaptations pile up. Both conditions
reach the same $\sim\!10^{-7}$ adaptation floor on every $D_k$, so the
retained $D_1$ behavior is not purchased with adaptation quality.

\begin{table}[t]
\centering
\small
\setlength{\tabcolsep}{4pt}
\renewcommand{\arraystretch}{1.15}
\begin{tabular}{lccc}
\toprule
\textbf{After} & \textbf{\textsc{No pause}} & \textbf{\textsc{Pause}} & \textbf{Gap} \\
\midrule
$D_2$ & $5.76 \pm 0.56$ & $2.74 \pm 0.81$ & $+3.01 \pm 0.75$ \\
$D_3$ & $8.60 \pm 1.44$ & $4.62 \pm 0.46$ & $+3.98 \pm 1.59$ \\
$D_4$ & $8.83 \pm 0.16$ & $6.04 \pm 0.84$ & $+2.80 \pm 0.97$ \\
\bottomrule
\end{tabular}
\caption{Held-out $D_1$ loss (nats, mean $\pm 1\sigma$, $n = 8$ seeds)
after each stage of the $D_1 \to D_2 \to D_3 \to D_4$ chain. Lower is
better; \textbf{Gap} is \textsc{No pause} $-$ \textsc{Pause}, so
positive values favor masked pauses.}
\label{tab:pilot_mlc_multistage}
\end{table}

\subsection{iGSM probing pilot}
\label{app:probe_pilot}

\paragraph{Base model and data.}
Qwen3-0.6B-Base, fine-tuned on $3{,}000$ problems sampled from the
iGSM generator~\cite{allenzhu2024igsm}. We use the
$\texttt{max\_op}{=}15$, $\texttt{max\_edge}{=}20$,
$\texttt{perm\_level}{=}5$ difficulty setting.

\paragraph{SFT.}
Three conditions, trained for $2$ epochs with learning rate
$1 \times 10^{-5}$, batch size $32$, AdamW optimizer, and bf16
precision: \textsc{No pause} (standard next-token prediction on
bare iGSM solutions) and \textsc{Pause} (one \texttt{[PAUSE]}
token inserted at every reasoning-step boundary, pause-target
positions excluded from the loss), and \textsc{Pause (unmasked)}, identical but with pause positions included in the loss. The \textsc{Base} condition is
the pretrained checkpoint with no fine-tuning. We report $n = 3$
seeds.

\paragraph{Probe training.}
For each held-out problem, we extract the final-layer hidden state
at the period token immediately preceding each reasoning step
boundary (the same character position across all four protocols).
We train two separate linear probes on these representations: one
predicting the integer answer of the \emph{current} step, one
predicting the integer answer of the \emph{next} step. Probes are
trained with a by-problem $70/30$ split, with each problem's
boundary positions assigned to either train or test based on the
problem split.

\section{Theoretical analysis for masked-pause SFT}
\label{app:pause_retention}
\subsection{A first-order view of masked-pause mode retention}

Old $D_1$ behaviours survive $D_2$ fine-tuning as long as their logits do
not move enough to erase their old margins. We make this precise on a
protected set of $D_1$ no-pause prefixes and then show that masked-pause
SFT yields a smaller certified protected logit drift than ordinary SFT.
The result is a first-order sufficient condition within the local
linearized model around $W_1$; it certifies preservation of the $D_1$ top
token on selected no-pause prefixes, not preservation of the full $D_1$
predictive distribution.

\paragraph{Margin stability.}
Whether an old top token survives depends only on whether its margin
outlasts the logit movement at that prefix; bounding the movement
uniformly over $\CC$ converts this into a retention guarantee.

\begin{lemma}[Margin stability]
\label{lem:margin}
Let $W_1$ be the shared parameters after $D_1$ training and $\CC$ a set
of $D_1$ no-pause prefixes whose top-token behaviour we protect. For
$c\in\CC$, let $a(c)=\arg\max_v z_{W_1}(c)_v$ be the $D_1$ top token and
$\gamma(c)$ its top-vs-runner-up margin at $W_1$ --- the amount a
competing logit must close to flip the prediction at $c$. Define the
margin survival function, the retention rate, and the protected logit
drift
\begin{align*}
\Phi(r)
&= \tfrac{1}{|\CC|}
   \bigl|\{c\in\CC:\gamma(c)>r\}\bigr|, \\
\Retain(W)
&= \tfrac{1}{|\CC|}
   \bigl|\{c\in\CC:
\\[-0.25em]
&\qquad \arg\max_v z_W(c)_v=a(c)\}\bigr|, \\
\Drift(W)
&= \max_{c\in\CC}\linfnorm{z_W(c)-z_{W_1}(c)} ,
\end{align*}
with $\Phi$ non-increasing in $r$. Then
\begin{equation*}
\Drift(W)\le\delta \;\Longrightarrow\; \Retain(W)\ge\Phi(2\delta).
\end{equation*}
\end{lemma}

The factor of two appears because the old top logit may move down by
$\delta$ while a competitor may move up by $\delta$.
Comparing the certified retention of two adaptation procedures
therefore reduces to comparing their certified protected logit drifts
on $\CC$.

\paragraph{What masked-pause SFT changes.}
Ordinary SFT predicts the next $D_2$ target $y_{i+1}$ from the no-pause
prefix $b_i$; masked-pause SFT predicts the same target from the
pause-perturbed prefix $b_i\circ p$ and masks the loss for the pause
token $p$ itself. Masking is essential: were $p$ a training target,
$b_i$ would again receive a direct supervised gradient, restoring the
interference channel masked-pause SFT removes. Four scalars control the
resulting first-order drifts on $\CC$. The interference exposure $\Iop$
is large when ordinary prefixes $b_i$ push the parameters in directions
that strongly move protected $D_1$ logits. Decomposing the Jacobian of
$b_i\circ p$ into a part inherited from the ordinary boundary Jacobian
and a pause-local residual yields a worst-case carry coefficient
$\lambda$ and a worst-case pause-local overlap $\nu$ with $\CC$.
Finally, $\kappa=M_P/M_O$ is the ratio of cumulative normalized
shared-update masses the two procedures need to reach the same $D_2$
fitting level (matched $D_2$ validation loss in our experiments).

\begin{lemma}[Protected drift bounds]
\label{lem:driftbounds}
In the local linearized regime around $W_1$ with $\Iop>0$, the
protected logit drifts of the ordinary and masked-pause endpoints
satisfy
\begin{align*}
\Drift(\Wo) &\le M_O\,\Iop, \\
\Drift(\Wp) &\le M_P\bigl(\lambda\Iop+\nu\bigr).
\end{align*}
Equivalently, defining $\bctx:=\kappa\bigl(\lambda+\nu/\Iop\bigr)$, the
masked-pause bound can be written as
\begin{equation*}
\Drift(\Wp) \;\le\; \bctx\,M_O\Iop.
\end{equation*}
\end{lemma}

Both bounds share the ordinary scale $M_O\Iop$, and in this certificate
$\bctx$ is the factor multiplying the masked-pause drift bound. Feeding these bounds
through Lemma~\ref{lem:margin} turns the comparison into a retention
guarantee.

\begin{theorem}[First-order context-perturbation retention certificate]
\label{thm:pause_retention}
Assume the local small-step regime around $W_1$, with ordinary and
masked-pause SFT using the same normal $D_2$ targets, reaching the same
$D_2$ fitting level, and $\Iop>0$. Combining the drift bounds of
Lemma~\ref{lem:driftbounds} with the margin-stability bound of
Lemma~\ref{lem:margin},
\begin{align*}
\Retain(\Wo) &\;\ge\; \Phi(2\,M_O\Iop), \\
\Retain(\Wp) &\;\ge\; \Phi(2\bctx\,M_O\Iop).
\end{align*}
Masked-pause SFT yields the stronger certificate whenever $\bctx<1$.
\end{theorem}

The condition $\bctx<1$ admits a geometric reading: masked-pause SFT
fits $D_2$ with less first-order drift on $D_1$ no-pause logits
whenever the pause state $b_i\circ p$ is sufficiently separated from
the ordinary boundary $b_i$ in the Jacobian sense captured by
$\lambda$ and $\nu$. This separation is structural rather than
architectural --- a property of the pause insertion policy, which our
empirical preference for reasoning-step boundaries effectively encodes.
The full setup, lemmas, proofs, and the role of masking and
pause-insertion position are in Appendix~\ref{app:pause_retention}.

\section{Theoretical analysis for masked-pause SFT - Proofs}
\label{app:pause_retention_proof}

This appendix contains the full setup, lemmas, and proofs supporting
Theorem~\ref{thm:pause_retention}. The analysis is a local first-order
sufficient-condition statement on a protected no-pause prefix set
$\CC$. It does \emph{not} claim preservation of the full next-token
distribution, of $D_1$ multimodality, or of the realized retained
fraction; it compares certified lower bounds within the local
linearized model around $W_1$.

\subsection{Setup and notation}
\label{app:setup}

Let $\VV$ be the vocabulary, $W$ the ordinary shared model parameters,
and $\ep$ the pause embedding. We write $\circ$ for token-sequence
concatenation, so $b_i\circ p$ denotes the prefix obtained by appending
a single pause token $p$ to $b_i$. For a no-pause prefix $c$, let
$z_W(c)\in\mathbb R^{|\VV|}$ be the next-token logit vector; for a
prefix containing $p$, write $z_{W,\ep}(s)$ when the dependence on $\ep$
should be explicit.

Let $W_1$ be the shared parameters after $D_1$ training and let $\CC$ be
the protected set of $D_1$ no-pause prefixes. For each $c\in\CC$, the
$D_1$ top token, its margin, and the margin survival function are
\begin{align*}
a(c) &= \arg\max_{v\in\VV} z_{W_1}(c)_v, \\
\gamma(c) &= z_{W_1}(c)_{a(c)} - \max_{v\neq a(c)} z_{W_1}(c)_v, \\
\Phi(r) &= \tfrac{1}{|\CC|}\bigl|\{c\in\CC:\gamma(c)>r\}\bigr|.
\end{align*}
Old-mode retention and protected no-pause logit drift are
\begin{align*}
\Retain(W)
&= \tfrac{1}{|\CC|}\!
   \sum_{c\in\CC}
   \mathbf{1}\!\bigl[
\\[-0.25em]
&\qquad \arg\max_{v}z_W(c)_v=a(c)
   \bigr], \\
\Drift(W)
&= \max_{c\in\CC}\linfnorm{z_W(c)-z_{W_1}(c)}.
\end{align*}
Because every prefix in $\CC$ is a no-pause prefix, both quantities
depend only on the shared parameters $W$; in particular, $\Drift(\Wp)$
and $\Retain(\Wp)$ do not directly involve the learned pause embedding
$\eph$.

For a matrix $B$, the operator norm induced by $\ell_\infty$ is
\begin{align*}
\opnorm{B}
&= \sup_{u\neq 0}\frac{\linfnorm{Bu}}{\linfnorm{u}} \\
&= \max_j\!\sum_k |B_{jk}|,
\end{align*}
and satisfies $\opnorm{XY}\le\opnorm{X}\opnorm{Y}$ whenever $XY$ is
defined.

\subsection{Direct supervised-prefix relocation}
\label{app:relocation}

Let $b_i=(x,y_1,\dots,y_i)$ be a no-pause $D_2$ reasoning boundary and
$\ell(z,y)$ the softmax cross-entropy loss for target $y$. The two
methods give the per-step direct losses
\begin{align*}
\ell_i^O(W) &= \ell\!\bigl(z_W(b_i),\,y_{i+1}\bigr), \\
\ell_i^P(W,\ep) &= \ell\!\bigl(z_{W,\ep}(b_i\circ p),\,y_{i+1}\bigr),
\end{align*}
with the pause-token prediction loss masked out in the second case.
Thus the direct supervised prefix for the normal target $y_{i+1}$ moves
from $b_i$ to $b_i\circ p$. Earlier tokens of $b_i$ can still receive
indirect gradient through the pause position; those contributions are
folded into the pause-state Jacobian below.

We denote the resulting (multi)sets of supervised prefixes by
\begin{align*}
\SO
&= \{b_i :
\\[-0.25em]
&\qquad \text{loss predicts a normal $D_2$ token}
\\[-0.25em]
&\qquad \text{from }b_i\}, \\
\SP
&= \{b_i\circ p :
\\[-0.25em]
&\qquad \text{loss predicts the same normal token}
\\[-0.25em]
&\qquad \text{from }b_i\circ p\},
\end{align*}
read as multisets if a prefix is supervised more than once.

\subsection{First-order interference and Jacobians}
\label{app:interference}

For a prefix $s$, define the start-of-adaptation shared-parameter logit
Jacobian
\begingroup
\renewcommand{\arraystretch}{1.45}
\begin{equation*}
J_1(s) =
\begin{cases}
\left.\dfrac{\partial z_W(s)}{\partial W}\right|_{W_1}
& s\ \text{no-pause}, \\
\left.\dfrac{\partial z_{W,\ep^{(0)}}(s)}{\partial W}\right|_{W_1}
& s\ \text{contains}\ p,
\end{cases}
\end{equation*}
\endgroup
where $\ep^{(0)}$ is the pause embedding at the start of $D_2$
adaptation; in the pause case, $\ep^{(0)}$ is held fixed and the
derivative is taken with respect to $W$ only. For $c\in\CC$ and a
supervised prefix $s$, the first-order interference matrix and exposure
are
\begin{align*}
K_1(c,s)
&= J_1(c) J_1(s)^\top, \\
\Ip(\CC,\SSet)
&= \max_{c\in\CC}
   \sum_{s\in\SSet}\opnorm{K_1(c,s)}.
\end{align*}
The operator-norm choice matches the $\ell_\infty$ logit-drift quantity
$\Drift$. We write $\Iop=\Ip(\CC,\SO)$. Operationally, $\Iop$ is large
when ordinary $D_2$ supervised prefixes share gradient directions with
protected $D_1$ prefixes.

\subsection{Pause-state Jacobian decomposition}
\label{app:decomp}

For each paired boundary, write $J_i^O=J_1(b_i)$ and
$J_i^P=J_1(b_i\circ p)$. Let $P_i^O$ be the orthogonal projection in
parameter space onto the row space of $J_i^O$, acting by
right-multiplication, so that $J_i^P P_i^O$ projects each row of $J_i^P$
onto $\mathrm{row}(J_i^O)$. Define
\begin{align*}
J_i^{\carry} &= J_i^P P_i^O, \\
J_i^{\loc}   &= J_i^P (I - P_i^O),
\end{align*}
so $J_i^P = J_i^{\carry} + J_i^{\loc}$. Because every row of
$J_i^{\carry}$ lies in $\mathrm{row}(J_i^O)$, there exists $A_i$ with
$J_i^{\carry}=A_i J_i^O$, and we take the canonical minimum-norm choice
\begin{equation*}
A_i = J_i^P (J_i^O)^+ ,
\end{equation*}
where $(\cdot)^+$ is the Moore--Penrose pseudoinverse. If $J_i^O=0$,
set $A_i=0$. The carry and pause-local interference summaries are
\begin{align*}
\lambda
&= \max_i \opnorm{A_i^\top}, \\
\nu
&= \max_{c\in\CC}
   \sum_i \opnorm{J_1(c)(J_i^{\loc})^\top}.
\end{align*}
Operationally, $\lambda$ is small when the pause state does not reuse
the same gradient-sensitive directions as the ordinary boundary, and
$\nu$ is small when the pause-local residual does not align with the
gradient-sensitive directions of protected $D_1$ prefixes. Whether
$\lambda$ and $\nu$ are small for a given architecture and insertion
policy is a separate structural question; see
Section~\ref{app:position}.

\subsection{Training-regime quantities}
\label{app:budget}

\begin{remark}[Local small-step regime]
The $D_2$ adaptation trajectory is analyzed in a neighbourhood of $W_1$
in which the linearization $z_W(s)\approx z_{W_1}(s)+J_1(s)(W-W_1)$ is
accurate; for the full nonlinear model, the same statements hold up to
a standard local linearization error in $\Drift$.
\end{remark}

For $A\in\{O,P\}$, let $w_t^A(s)$ be the normalized weight of supervised
prefix $s$ in the minibatch loss at step $t$ and $\eta_t$ the learning
rate. Define
\begin{align*}
m_A(s) &= \sum_t \eta_t w_t^A(s), \\
M_A &= \max_s m_A(s), \qquad \kappa = \frac{M_P}{M_O}.
\end{align*}
$M_A$ is the largest cumulative normalized update mass assigned to any
single supervised prefix along the trajectory. Throughout this
appendix, ``the same $D_2$ fitting level'' refers to matched $D_2$
validation loss (or task accuracy, when that is the empirical
comparison metric used in the main paper); $\kappa>1$ then quantifies
the additional shared-parameter update mass masked-pause SFT requires,
if any, to reach that level.

\subsection{Additional lemmas}
\label{app:lemmas}

(Lemmas~\ref{lem:margin} and~\ref{lem:driftbounds} are stated in the
main text; their proofs are included below.)

\begin{lemma}[Direct-loss relocation]
\label{lem:reloc}
Write $J_W(s)=\partial z_W(s)/\partial W$ and
$J_{W\mid\ep}(s)=\partial z_{W,\ep}(s)/\partial W$ (with $\ep$ held
fixed). Then
\begin{align*}
\nabla_W\ell_i^O(W)
&= J_W(b_i)^\top r_i^O, \\[0.35em]
\nabla_W\ell_i^P(W,\ep)
&= J_{W\mid\ep}(b_i\circ p)^\top r_i^P,
\end{align*}
with $r_i^O=\nabla_z\ell(z_W(b_i),y_{i+1})$ and
$r_i^P=\nabla_z\ell(z_{W,\ep}(b_i\circ p),y_{i+1})$. There is no direct
loss term of the form $\ell(z_W(b_i),p)$, because the pause-token
prediction loss is masked. Under the local-linearized convention,
$J_W$ and $J_{W\mid\ep}$ reduce to $J_1$ at the start of $D_2$
adaptation.
\end{lemma}

\begin{lemma}[Interference separation]
\label{lem:sep}
Under the canonical decomposition of Section~\ref{app:decomp},
\begin{equation*}
\Ip(\CC,\SP) \;\le\; \lambda\,\Iop + \nu.
\end{equation*}
If $\Iop>0$, equivalently
$\Ip(\CC,\SP)\le(\lambda+\nu/\Iop)\,\Iop=\actx\,\Iop$.
\end{lemma}

\begin{lemma}[Interference controls drift]
\label{lem:drift}
In the local linearized regime, any $D_2$ adaptation procedure with
supervised multiset $\SSet$ and normalized update masses $m(s)$ satisfies
\begin{align*}
\Drift(W)
&\le \max_{c\in\CC}
    \sum_{s\in\SSet} m(s)\opnorm{K_1(c,s)} \\
&\le M\,\Ip(\CC,\SSet),
\end{align*}
where $M=\max_s m(s)$.
\end{lemma}

\subsection{Proofs}
\label{app:proof}

\paragraph{Proof of Lemma~\ref{lem:margin} (margin stability).}
Fix $c\in\CC$ with $\gamma(c)>2\delta$ and any $v\neq a(c)$. Since each
logit moves by at most $\delta$,
\begin{align*}
&z_W(c)_{a(c)}-z_W(c)_v \\
&\quad\ge \bigl(z_{W_1}(c)_{a(c)}-z_{W_1}(c)_v\bigr)-2\delta \\
&\quad\ge \gamma(c)-2\delta \;>\; 0.
\end{align*}
Thus $a(c)$ remains the top token under $W$, and the fraction of $c$
with $\gamma(c)>2\delta$ lower-bounds $\Retain(W)$. $\square$

\paragraph{Proof of Lemma~\ref{lem:reloc}.}
Ordinary SFT predicts $y_{i+1}$ from $b_i$, so its CE gradient is
attached to the output state at $b_i$; the chain rule gives the stated
form. Masked-pause SFT inserts $p$ after $b_i$, masks the would-be loss
for predicting $p$ from $b_i$, and predicts $y_{i+1}$ from the
pause-state output position $b_i\circ p$; again the chain rule gives
the stated form. Indirect gradient on earlier tokens flows through
$J_{W\mid\ep}(b_i\circ p)$ and is handled by the carry/local
decomposition. $\square$

\paragraph{Proof of Lemma~\ref{lem:sep}.}
Fix $c\in\CC$. Using $J_i^P=A_i J_i^O+J_i^{\loc}$,
\begin{align*}
K_1(c,b_i\circ p)
&= J_1(c)(J_i^P)^\top \\
&= K_1(c,b_i)\,A_i^\top + J_1(c)(J_i^{\loc})^\top.
\end{align*}
By submultiplicativity of $\opnorm{\cdot}$,
\begin{align*}
\opnorm{K_1(c,b_i\circ p)}
&\le \lambda\opnorm{K_1(c,b_i)} \\
&\quad + \opnorm{J_1(c)(J_i^{\loc})^\top}.
\end{align*}
Summing over $i$ and taking $\max_{c\in\CC}$ on both sides, with
$\max(A(c)+B(c))\le\max A(c)+\max B(c)$, gives
$\Ip(\CC,\SP)\le\lambda\Iop+\nu$. $\square$

\paragraph{Proof of Lemma~\ref{lem:drift}.}
In the linearized regime, the step is
$\Delta W_t=-\eta_t\sum_{s}w_t(s)J_1(s)^\top r_t(s)$, where the softmax
CE residual $r_t(s)=q_t(s)-e_{y_s}$ satisfies $\linfnorm{r_t(s)}\le1$
uniformly along the trajectory. For $c\in\CC$, summing first-order
logit changes,
\begin{align*}
\linfnorm{\Delta z(c)}
&\le \sum_t\eta_t\sum_{s}w_t(s) \\
&\qquad \opnorm{K_1(c,s)}\linfnorm{r_t(s)} \\
&\le \sum_{s} m(s)\opnorm{K_1(c,s)}.
\end{align*}
Taking $\max_{c\in\CC}$ and then $M=\max_s m(s)$ gives both
inequalities. $\square$

\paragraph{Proof of Lemma~\ref{lem:driftbounds}.}
Applying Lemma~\ref{lem:drift} to the ordinary supervised multiset
$\SO$ gives $\Drift(\Wo)\le M_O\,\Ip(\CC,\SO)=M_O\Iop$. For masked-pause
SFT, Lemma~\ref{lem:drift} on $\SP$ followed by Lemma~\ref{lem:sep}
gives
\begin{align*}
\Drift(\Wp)
&\le M_P\,\Ip(\CC,\SP) \\
&\le M_P\bigl(\lambda\Iop+\nu\bigr).
\end{align*}
When $\Iop>0$, factoring $M_O\Iop$ out of the last bound gives
\begin{align*}
M_P\bigl(\lambda\Iop+\nu\bigr)
&= \frac{M_P}{M_O}\!
   \left(\lambda+\frac{\nu}{\Iop}\right) M_O\Iop \\
&= \bctx\,M_O\Iop. \qquad\square
\end{align*}

\paragraph{Proof of Theorem~\ref{thm:pause_retention}.}
By Lemma~\ref{lem:driftbounds}, $\Drift(\Wo)\le M_O\Iop$ and
$\Drift(\Wp)\le\bctx\,M_O\Iop$. Applying Lemma~\ref{lem:margin} with
$\delta=M_O\Iop$ and $\delta=\bctx\,M_O\Iop$ respectively yields
\begin{align*}
\Retain(\Wo) &\ge \Phi(2\,M_O\Iop), \\
\Retain(\Wp) &\ge \Phi(2\bctx\,M_O\Iop).
\end{align*}
Since $\Phi$ is non-increasing and $\Iop>0$, the condition $\bctx<1$
gives $\bctx\,M_O\Iop<M_O\Iop$, so the masked-pause certificate
dominates the ordinary-SFT one. The dominance is strict whenever a
nonzero fraction of $\CC$ has
$\gamma(c)\in(2\bctx\,M_O\Iop,\,2\,M_O\Iop]$. $\square$

\subsection{Why masking the pause-token loss matters}
\label{app:masking}

If the pause-token prediction loss is not masked, each inserted pause
introduces an extra direct loss
$\ell_i^{\mathrm{pause}}(W)=\ell(z_W(b_i),p)$ whose gradient is attached
to the original no-pause boundary $b_i$. This re-introduces a
supervised multiset
$\mathcal S_{\mathrm{pause}}=\{b_i:\text{loss predicts $p$ from }b_i\}$
and changes the drift certificate to
\begin{align*}
\Drift(\Wp^{\mathrm{unmask}})
&\le M_{\mathrm{norm}}(\lambda\Iop+\nu) \\
&\quad + M_{\mathrm{pause}}\,I_{\mathrm{pause}},
\end{align*}
where $I_{\mathrm{pause}}=\Ip(\CC,\mathcal S_{\mathrm{pause}})$. Since
$\mathcal S_{\mathrm{pause}}$ consists of the original no-pause
boundaries, this added term can erase the benefit of context
perturbation. Masking removes this loss and leaves
Theorem~\ref{thm:pause_retention} clean.

\subsection{Pause insertion position}
\label{app:position}

An insertion policy $A$ determines $\SP^A$ and therefore changes
$\lambda(A)$, $\nu(A)$, and $\kappa(A)=M_P^A/M_O$, giving
$\bctx(A)=\kappa(A)(\lambda(A)+\nu(A)/\Iop)$. The theorem applies as
soon as $\bctx(A)<1$. The remark below collects the qualitative
considerations; it is not part of the formal theorem.

\begin{remark}[Why reasoning-step boundaries are natural insertion points]
\label{rem:position}
(i) At a reasoning-step boundary the previous step has ended, so the
pause state $b_i\circ p$ is less pressured to copy a mid-phrase token
representation, which tends to reduce the carry coefficient
$\lambda(A)$. (ii) The pause-local residual at a complete boundary
need not preserve ordinary syntactic-continuation features tied to
$D_1$, which tends to reduce $\nu(A)$. (iii) Predicting the next
reasoning step from a boundary-plus-pause prefix is still a natural
next-token task, so the fitting-budget ratio $\kappa(A)$ stays
moderate. Insertion that is too sparse leaves most supervised prefixes
close to $\SO$; insertion at unnatural dense positions can force the
pause state to copy the immediately preceding prefix, increasing
$\lambda(A)$, $\nu(A)$, or $\kappa(A)$.
\end{remark}

\section{Experimental Settings}
\label{app:experimental_settings}

\subsection{Hyperparameter settings}
\label{app:hyperparameters}
We report the hyperparameter settings used for each training stage. Across experiments, we use FusedAdam as the optimizer, CosineAnnealingLR as the learning-rate scheduler, a warmup ratio of 0.03, and bf16 precision for training, while adapting the learning rate and batch configuration to each training objective and model family.

For SFT on mathematical reasoning data, we train each model for 3 epochs with a batch size of 32. The learning rate is selected depending on the model family: \(5\times 10^{-6}\) for Qwen models, \(1\times 10^{-5}\) for Llama3.2, and \(5\times 10^{-5}\) for Llama3.1. The full set of hyperparameters for math SFT is provided in Table~\ref{tab:sft_math_hparams}.

For SFT on code generation data, we also train for 3 epochs with a batch size of 32, using the same optimizer, scheduler, warmup ratio, and training precision as in math SFT. The learning rate is model-specific: \(5\times 10^{-7}\) for Qwen3-1.7B-Base, \(5\times 10^{-6}\) for Qwen3-4B-Base, \(5\times 10^{-5}\) for Qwen3-8B-Base, and \(1\times 10^{-8}\) for Llama3.1-8B-Instruct. The complete configuration for code SFT is summarized in Table~\ref{tab:sft_code_hparams}.

For GRPO training, we use a learning rate of \(5\times 10^{-7}\), train for 1 episode, and use a batch size of 128. For each prompt, we sample 8 rollouts and apply a KL coefficient of 0.001. The optimizer, scheduler, warmup ratio, and training dtype are kept consistent with the SFT settings. The GRPO hyperparameters are listed in Table~\ref{tab:grpo_hparams}.
\begin{table}[t]
\centering
\small
\setlength{\tabcolsep}{4pt}
\renewcommand{\arraystretch}{1.15}
\begin{tabular}{p{0.36\columnwidth}p{0.52\columnwidth}}
\toprule
\textbf{Hyperparameter} & \textbf{SFT on math} \\
\midrule
Learning rate
& Qwen: \(5\times 10^{-6}\);
  Llama3.2: \(1\times 10^{-5}\);
  Llama3.1: \(5\times 10^{-5}\) \\
Epochs & 3 \\
Batch size & 32 \\
Optimizer & FusedAdam \\
Scheduler & CosineAnnealingLR \\
Warmup ratio & 0.03 \\
Training dtype & bf16 \\
\bottomrule
\end{tabular}
\caption{SFT hyperparameters for mathematical reasoning training.}
\label{tab:sft_math_hparams}
\end{table}

\begin{table}[t]
\centering
\small
\setlength{\tabcolsep}{4pt}
\renewcommand{\arraystretch}{1.15}
\begin{tabular}{p{0.34\columnwidth}p{0.54\columnwidth}}
\toprule
\textbf{Hyperparameter} & \textbf{SFT on code} \\
\midrule
Learning rate
& Qwen3-1.7B-Base: \(5\times 10^{-7}\);
  Qwen3-4B-Base: \(5\times 10^{-6}\);
  Qwen3-8B-Base: \(5\times 10^{-5}\);
  Llama3.1-8B-Instruct: \(1\times 10^{-8}\) \\
Epochs & 3 \\
Batch size & 32 \\
Optimizer & FusedAdam \\
Scheduler & CosineAnnealingLR \\
Warmup ratio & 0.03 \\
Training dtype & bf16 \\
\bottomrule
\end{tabular}
\caption{SFT hyperparameters for code generation training.}
\label{tab:sft_code_hparams}
\end{table}

\begin{table}[t]
\centering
\small
\setlength{\tabcolsep}{4pt}
\renewcommand{\arraystretch}{1.15}
\begin{tabular}{p{0.42\columnwidth}p{0.46\columnwidth}}
\toprule
\textbf{Hyperparameter} & \textbf{GRPO} \\
\midrule
Learning rate & \(5\times 10^{-7}\) \\
Episodes & 1 \\
Batch size & 128 \\
Rollouts per prompt & 8 \\
Optimizer & FusedAdam \\
Scheduler & CosineAnnealingLR \\
Warmup ratio & 0.03 \\
Training dtype & bf16 \\
KL coefficient & 0.001 \\
\bottomrule
\end{tabular}
\caption{GRPO hyperparameters.}
\label{tab:grpo_hparams}
\end{table}

\subsection{Computation resources}
\label{app:compute}

All training experiments in this paper were conducted with OpenRLHF\cite{hu2024openrlhf} on
2$\times$ NVIDIA RTX PRO 6000 and 2$\times$ NVIDIA RTX A100 GPUs.

\subsection{Dataset Statistics}
\label{app:datsaet_statistics}

We used a mixture of data from GSM8K, MATH, and DeepMATH~\cite{cobbe2021training, hendrycks2021measuring, deepmath103k} for math domain. Table~\ref{tab:pause_dataset_stats} summarizes the statistics of the mathematical datasets used for training and evaluation.

\begin{table}[t]
\centering
\small
\begin{tabular}{lrrr}
\toprule
Statistic & Train & Eval & Total \\
\midrule
Examples                 & 83,438 & 4,391 & 87,829 \\
Full SFT tokens          & 44.14M & 2.36M & 46.51M \\
Pause tokens             & 1.34M  & 71.72K & 1.41M \\
Pause-token ratio        & 3.02\% & 3.03\% & 3.03\% \\
Pauses/sample            & 16.003 & 16.332 & 16.020 \\
\bottomrule
\end{tabular}
\caption{Statistics of the step-aligned pause dataset.}
\label{tab:pause_dataset_stats}
\end{table}

\subsection{Evaluation setup}
\label{app:evaluation_setup}

For mathematical reasoning benchmarks, we report pass@1 accuracy under
a deterministic decoding setting with temperature 0.0. For code
generation benchmarks, we report pass@1 using the EvalPlus package, and
a prediction is counted as correct only when it passes all test cases.
For general benchmarks, we report pass@4 accuracy with temperature 0.7
using the LightEval package.

\section{Implementation Details}
\label{app:implementation_details}
We implemented most components using OpenRLHF\cite{hu2024openrlhf}, and used the default OpenRLHF settings for hyperparameters not explicitly specified.

\subsection{Objective function}
\label{app:objective_function}

Let \(x\) denote the prompt and
\(y=(y_1,\ldots,y_T)\) denote the target response tokens. For a prefix
\(s\), write \(p_\theta(\cdot\mid s)\) for the model distribution and
\(\ell_\theta(s,a)=-\log p_\theta(a\mid s)\) for the next-token
cross-entropy loss. Standard SFT optimizes
\begin{equation*}
\mathcal L_{\mathrm{SFT}}
= \frac{1}{T}\sum_{t=1}^{T}
  \ell_\theta(x\circ y_{<t}, y_t).
\end{equation*}

For pause-based training, let
\(u=(u_1,\ldots,u_L)\) be the sequence obtained after inserting pause
tokens into \(y\), where each \(u_j\) is either a normal target token or
the pause token \(p\). Masked-pause SFT uses the same next-token CE
loss, but removes the loss terms whose target is \(p\):
\begin{align*}
m_j &= \mathbf 1[u_j\neq p], \\
\mathcal L_{\mathrm{mask}}
&= \frac{1}{\sum_j m_j}
   \sum_{j=1}^{L}
   m_j\,\ell_\theta(x\circ u_{<j}, u_j).
\end{align*}
Thus the model is trained to predict normal answer tokens from prefixes
that may contain pauses, while it is not directly trained to emit the
pause token itself. The pause-predicted variant keeps the same inserted
sequence \(u\), but does not mask pause targets:
\begin{equation*}
\mathcal L_{\mathrm{pred}}
= \frac{1}{L}\sum_{j=1}^{L}
  \ell_\theta(x\circ u_{<j}, u_j).
\end{equation*}
The distinction is important: masked-pause SFT changes the context used
for predicting ordinary tokens, whereas pause-predicted SFT additionally
adds direct supervision for predicting \(p\).

\subsection{Step-Boundary Extraction}
\label{app:step_boundary_extraction}

We identify reasoning-step boundaries with deterministic preprocessing.
For math, line breaks are the primary boundary candidates, while wrapped
continuations and line breaks inside inline or display-math expressions
are preserved. Displayed equations written with \verb|\[...\]|,
\verb|$$...$$|, or standard equation and alignment environments remain
intact and are attached to immediately preceding explanatory text when
present. Outside math environments, we split at transitions that form an
independent reasoning unit, including bullets, final-answer headings,
discourse cues, and completed calculations, and merge short
non-semantic fragments with the preceding step. For code, we retain the
extracted code region, discard surrounding natural-language text and
blank lines, and treat each remaining source line as one step. One pause
token is inserted at each resulting boundary, and its target position is
masked from the loss.

\subsection{Training Overhead}
\label{app:training_overhead}

MBP inserts approximately 16 pause tokens per math example and 27 per
code example on average. Pause tokens account for approximately $2.2\%$
and $9.8\%$ of tokens in the corresponding augmented math and code
corpora, respectively. Training is otherwise identical to standard SFT:
pause positions are simply excluded from the loss, and insertion is a
deterministic rule-based preprocessing step.

\subsection{MI-Proxy Computation}
\label{app:mi_proxy_computation}

For each prompt $q_i$, we take $y_i$ to be the final balanced expression in the generated solution and let $x_{i,k}$
denote the cumulative solution prefix ending at the $k$-th step boundary
before the answer. For model $m$, we compute
\begin{equation*}
r_{i,k}^{(m)}
=
\log
\frac{p_m(x_{i,k},y_i\mid q_i)}
     {p_m(x_{i,k}\mid q_i)\,p_m(y_i\mid q_i)}.
\end{equation*}
We weight each observed pair by its joint sequence probability,
$w_{i,k}^{(m)}=p_m(x_{i,k},y_i\mid q_i)r_{i,k}^{(m)}$, and macro-average
first over the $K_i$ boundaries of each example and then over the $N$
examples:
\begin{equation*}
\widehat I_w(m)
=
\frac{1}{N}
\sum_{i=1}^{N}
\frac{1}{K_i}
\sum_{k=1}^{K_i}
w_{i,k}^{(m)}.
\end{equation*}
We report the signed, Base-referenced normalized score
\begin{align*}
\Delta_B(m) &= \widehat I_w(m)-\widehat I_w(B), \\
G_B(m)
&= \operatorname{sign}\!\left(\Delta_B(m)\right) \\
&\quad \cdot \log_{10}\!\left(
1+\frac{|\Delta_B(m)|}{|\widehat I_w(B)|}
\right),
\end{align*}
with $G_B(B)=0$, where $B$ denotes the Base model. All results use
macro aggregation over the full evaluation set
(\texttt{filter = all}).

\subsection{Examples of Methods}
\label{app:method_examples}

The following examples illustrate the actual data formats used for different pause-token placement strategies, including MBP (Masked Boundary Pause), Random, Append, and DIT. Examples for math data are shown in Table~\ref{tab:math_method_examples}, while examples for code data are provided in Table~\ref{tab:code_method_examples} and Table~\ref{tab:code_method_examples_cont}.

\begin{table*}[p]
\centering
\scriptsize
\setlength{\tabcolsep}{4pt}
\renewcommand{\arraystretch}{1.15}
\begin{tabular}{p{0.13\textwidth}p{0.80\textwidth}}
\toprule
\textbf{Method} & \textbf{Math example} \\
\midrule
MBP
& \textbf{Question.} Round each number to the nearest tenth and add the
results: 543.55, 25.23, and 299.98.
\par\textbf{Answer.} \pausetok{}First, let's round each number to the
nearest tenth: \pausetok{}1. 543.55 rounds to 543.6 (since the
hundredths place is 5). \pausetok{}2. 25.23 rounds to 25.2 (since the
hundredths place is less than 5). \pausetok{}3. 299.98 rounds to 300.0
(since the hundredths place is 8). \pausetok{}Now, let's add the
rounded numbers: \pausetok{}543.6 + 25.2 + 300.0 = 868.8.
\pausetok{}So, the final answer is \(\boxed{868.8}\). \\
\midrule
append
& \textbf{Question.} Round each number to the nearest tenth and add the
results: 543.55, 25.23, and 299.98.\pausetok{}\pausetok{}\pausetok{}
\pausetok{}\pausetok{}\pausetok{}\pausetok{}
\par\textbf{Answer.} First, let's round each number to the nearest
tenth: 1. 543.55 rounds to 543.6 (since the hundredths place is 5).
2. 25.23 rounds to 25.2 (since the hundredths place is less than 5).
3. 299.98 rounds to 300.0 (since the hundredths place is 8). Now, let's
add the rounded numbers: 543.6 + 25.2 + 300.0 = 868.8. So, the final
answer is \(\boxed{868.8}\). \\
\midrule
random
& \textbf{Question.} Round each number to the nearest tenth and add the
results: 543.55, 25.23, and 299.98.
\par\textbf{Answer.} First, let's round each number to the nearest
tenth: 1. 543.55 rounds to 5\pausetok{}4\pausetok{}3.6 (since the
hundredths place is 5). 2. 25.23 rounds to 25.\pausetok{}2 (since the
hundred\pausetok{}ths place is less than 5). 3. 299.98 rounds to
3\pausetok{}00.0 (since the hundredths place is 8). Now, let's add the
rounded numbers: 543.6 + 25.2 + 300.0 = 868.8. So, the
final\pausetok{} answer is \(\backslash\)boxed\pausetok{}\{868.8\}. \\
\midrule
dit
& \textbf{Question.} Round each number to the nearest tenth and add the
results: 543.55, 25.23, and 299.98.
\par\textbf{Answer.} \pausetok{}First,\pausetok{} let's round each
number to the nearest tenth: \pausetok{}1. 543.55 rounds to 543.6
(since the hundredths place is 5\pausetok{}). 2. 25.23 rounds to 25.2
(since the hundredths place is\pausetok{} less than 5). 3. 299.98
rounds to 300.0 (since the hundredths place is 8). Now, let's add the
rounded numbers: 543.6 + 25.2 + 300.0 = 868.8. So, the\pausetok{}
final answer is\pausetok{} \(\boxed{868.8}\). \\
\bottomrule
\end{tabular}
\caption{Examples of pause-token placement for the math task.}
\label{tab:math_method_examples}
\end{table*}

\providecommand{\appexcodequestion}{Ninja Atsushi stands guard atop the
Ninja Building, which consists of two adjacent buildings of equal
height. Each building has a series of walls of varying heights. Atsushi
must jump between buildings to patrol the rooftops. Given the heights of
the walls for both buildings, determine the minimum height Atsushi must
jump for each wall position. Input: two lists of integers,
\texttt{heights\_building\_1} and \texttt{heights\_building\_2}, with
the same length. Output: a list of integers where each element is the
minimum height Atsushi must jump for each wall position. Constraints:
the common list length is between 1 and \(10^5\), and each wall height
is between 1 and \(10^9\). Sample input:
\texttt{heights\_building\_1 = [3, 6, 2]},
\texttt{heights\_building\_2 = [5, 4, 8]}. Sample output:
\texttt{[5, 6, 8]}.}

\newcommand{\codeindent}{\hspace*{1.6em}}
\newcommand{\codeblock}[1]{%
  \par\smallskip
  \begin{minipage}[t]{0.98\linewidth}
  \ttfamily\raggedright\setlength{\parindent}{0pt}
  #1
  \end{minipage}%
}

\begin{table*}[p]
\centering
\scriptsize
\setlength{\tabcolsep}{4pt}
\renewcommand{\arraystretch}{1.15}
\begin{tabular}{p{0.13\textwidth}p{0.80\textwidth}}
\toprule
\textbf{Method} & \textbf{Code example} \\
\midrule
MBP
&
\textbf{Question.} \appexcodequestion
\par\textbf{Answer.}
\codeblock{
\pausetok{}def min\_jumps(heights\_building\_1,\\
\codeindent heights\_building\_2):\\
\codeindent \pausetok{}Calculate the minimum height Atsushi must jump\\
\codeindent for each wall position between two buildings.\\
\codeindent \pausetok{}:param heights\_building\_1: list of integers representing\\
\codeindent the heights of walls in the first building.\\
\codeindent \pausetok{}:param heights\_building\_2: list of integers representing\\
\codeindent the heights of walls in the second building.\\
\codeindent \pausetok{}:return: list of integers representing\\
\codeindent the minimum jump heights for each wall position.\\
\codeindent \pausetok{}if len(heights\_building\_1) != len(heights\_building\_2):\\
\codeindent\codeindent \pausetok{}raise ValueError("Both buildings must have the same number of walls.").\\
\codeindent \pausetok{}return [\\
\codeindent\codeindent max(heights\_building\_1[i], heights\_building\_2[i])\\
\codeindent\codeindent for i in range(len(heights\_building\_1))\\
\codeindent ].
}
\\
\midrule
append
&
\textbf{Question.} \appexcodequestion\pausetok{}\pausetok{}
\pausetok{}\pausetok{}\pausetok{}\pausetok{}\pausetok{}\pausetok{}
\pausetok{}\pausetok{}
\par\textbf{Answer.}
\codeblock{
def min\_jumps(heights\_building\_1,\\
\codeindent heights\_building\_2):\\
\codeindent Calculate the minimum height Atsushi must jump\\
\codeindent for each wall position between two buildings.\\
\codeindent :param heights\_building\_1: list of integers representing\\
\codeindent the heights of walls in the first building.\\
\codeindent :param heights\_building\_2: list of integers representing\\
\codeindent the heights of walls in the second building.\\
\codeindent :return: list of integers representing\\
\codeindent the minimum jump heights for each wall position.\\
\codeindent if len(heights\_building\_1) != len(heights\_building\_2):\\
\codeindent\codeindent raise ValueError("Both buildings must have the same number of walls.").\\
\codeindent return [\\
\codeindent\codeindent max(heights\_building\_1[i], heights\_building\_2[i])\\
\codeindent\codeindent for i in range(len(heights\_building\_1))\\
\codeindent ].
}
\\
\bottomrule
\end{tabular}
\caption{Examples of pause-token placement for the code task.}
\label{tab:code_method_examples}
\end{table*}

\begin{table*}[p]
\centering
\scriptsize
\setlength{\tabcolsep}{4pt}
\renewcommand{\arraystretch}{1.15}
\begin{tabular}{p{0.13\textwidth}p{0.80\textwidth}}
\toprule
\textbf{Method} & \textbf{Code example} \\
\midrule
random
&
\textbf{Question.} \appexcodequestion
\par\textbf{Answer.}
\codeblock{
def min\_jumps(heights\_building\pausetok{}\_1,\\
\codeindent \pausetok{} heights\_building\_2):\\
\codeindent Calculate the minimum height Atsushi must jump\\
\codeindent for each wall position between two buildings.\\
\codeindent :param heights\_building\_1: list of integers representing\\
\codeindent the heights of walls in the first building.\\
\codeindent \pausetok{}:param heights\_building\_2: list of integers representing\\
\codeindent the heights of walls in the second building.\\
\codeindent :return: list of integers representing\\
\codeindent the minimum jump heights for each wall position\pausetok{}.\\
\codeindent if len\pausetok{}(height\pausetok{}s\_building\_1) != len(heights\_building\_2):\\
\codeindent\codeindent raise ValueError("Both buildings must have the same number of walls.").\\
\codeindent return [\\
\codeindent\codeindent max(heights\_building\_1[i], heights\_building\_2\pausetok{}[i\pausetok{}])\\
\codeindent\codeindent for i in range\pausetok{}(len(heights\_building\_1\pausetok{}))\\
\codeindent ].
}
\\
\midrule
dit
&
\textbf{Question.} \appexcodequestion
\par\textbf{Answer.}
\codeblock{
\pausetok{}def min\pausetok{}\_jumps(\\
\codeindent heights\_building\_1, heights\_building\_2):\\
\codeindent Calculate the minimum height Atsushi must jump\\
\codeindent for each wall position\pausetok{} between two buildings.\\
\codeindent \pausetok{}:param heights\_building\_1: list of integers representing\\
\codeindent the heights of\pausetok{} walls in the first building.\\
\codeindent :param heights\_building\_2: list of integers representing\\
\codeindent the heights of walls in the second building.\\
\codeindent :return: list of integers representing the minimum\pausetok{}\\
\codeindent jump heights for each wall position.\\
\codeindent \pausetok{}if len(heights\_building\_1) != len(heights\_building\_2):\\
\codeindent\codeindent raise ValueError("Both buildings must have the same number of walls\pausetok{}.").\\
\codeindent \pausetok{}return [\\
\codeindent\codeindent max(heights\_building\_1[i], heights\_building\_2[i])\\
\codeindent\codeindent for i in range(len(heights\_building\_1))\\
\codeindent ].\pausetok{}
}
\\
\bottomrule
\end{tabular}
\caption{Examples of pause-token placement for the code task, continued.}
\label{tab:code_method_examples_cont}
\end{table*}

\section{Additional Results}
\label{app:additional_results}

This section reports the full evaluation results that are not shown in the main text. We include mathematical reasoning, code generation,
and general benchmark results for Qwen3-1.7B-Base, Qwen3-4B-Base,
Qwen3-8B-Base, Llama3.2-1B-Instruct, Llama3.2-3B-Instruct, and
Llama3.1-8B-Instruct. The tables compare the base models, standard SFT,
and the pause-insertion variants Append, Random,
DIT, MBP, and MBP w/o masking where available.
The full results for math and general language understanding tasks are shown in Table~\ref{tab:main_math_results}, while the full results for code and general language understanding tasks are presented in Table~\ref{tab:main_code_results}. Table~\ref{tab:grpo_accuracy_auc_full} reports the full six-benchmark GRPO results.

\begin{table*}[t]
    \centering
    \small
    \setlength{\tabcolsep}{6pt}
    \renewcommand{\arraystretch}{1.15}
    \resizebox{\textwidth}{!}{%
    \begin{tabular}{lcccccccccccc}
    \toprule
    \textbf{Method}
    & \textbf{GSM8K}
    & \textbf{MATH500}
    & \textbf{MATH}
    & \textbf{AIME24}
    & \textbf{AMC23}
    & \textbf{MINERVA}
    & \textbf{MMLU}
    & \textbf{GPQA}
    & \textbf{HellaSwag}
    & \textbf{BBH}
    & \textbf{PIQA}
    & \textbf{Average} \\
    
    \modelrow{13}{Qwen3-1.7B-Base}
    Base                            & 65.13 & 54.20 & 54.82 & 0.00 & \underline{35.00} & 17.65 & 62.95 & \textbf{27.46} & 38.69 & 38.84 & 62.13 & 41.53 \\
    SFT                             & \underline{82.41} & \textbf{63.40} & 62.98 & 0.00 & 27.50 & 20.59 & 62.32 & 24.78 & 40.39 & 42.06 & 66.10 & 44.78 \\
    Append                          & 79.68 & 59.20 & 60.96 & 0.00 & 30.00 & \underline{22.79} & 62.93 & 25.67 & 41.53 & 42.62 & 66.59 & 44.72 \\
    Random                          & 81.35 & 60.80 & \underline{63.26} & 0.00 & 20.00 & \textbf{23.16} & \underline{63.64} & 23.66 & \textbf{48.73} & 43.96 & \textbf{69.64} & 45.29 \\
    DIT                             & 68.01 & 57.20 & 56.38 & \textbf{6.67} & 30.00 & 20.22 & 62.90 & 25.67 & 44.75 & 42.66 & \underline{69.31} & 43.98 \\
    \oursrule{13}
    \textbf{MBP}                    & \textbf{83.02} & 61.00 & \textbf{64.56} & \underline{3.33} & \textbf{37.50} & \textbf{23.16} & \textbf{63.93} & \underline{27.23} & \underline{48.32} & \underline{44.13} & 67.63 & \textbf{47.62} \\
    \textbf{MBP w/o masking}        & 80.97 & \underline{62.20} & 62.64 & \textbf{6.67} & 27.50 & \textbf{23.16} & 63.53 & 24.55 & 45.86 & \textbf{44.17} & 66.70 & \underline{46.18} \\
    
    \modelrow{13}{Qwen3-4B-Base}
    Base                            & 82.56 & 66.00 & 67.64 & \underline{6.67} & 37.50 & 28.68 & 73.05 & \textbf{30.80} & 63.69 & \textbf{56.64} & 73.29 & 53.32 \\
    SFT                             & \underline{91.36} & 69.20 & \underline{71.24} & \underline{6.67} & \underline{45.00} & \underline{29.78} & 73.33 & 29.91 & 75.53 & 54.64 & 77.91 & 56.78 \\
    Append                          & 89.92 & 67.00 & 68.62 & \textbf{10.00} & 42.50 & 28.68 & 73.35 & 29.02 & \textbf{76.47} & \underline{54.94} & 78.45 & 56.27 \\
    Random                          & \textbf{91.43} & \underline{70.40} & 71.06 & \textbf{10.00} & 42.50 & 29.04 & \underline{73.57} & 28.79 & \underline{76.14} & 54.89 & 77.64 & \underline{56.86} \\
    DIT                             & 87.19 & 63.00 & 64.00 & \underline{6.67} & 40.00 & 29.04 & \textbf{73.60} & \underline{30.36} & 70.35 & 54.49 & \underline{79.43} & 54.38 \\
    \oursrule{13}
    \textbf{MBP}                    & 90.07 & \textbf{73.40} & \textbf{72.32} & \textbf{10.00} & \textbf{50.00} & \textbf{30.51} & 73.54 & 26.79 & 76.04 & 54.55 & \textbf{79.65} & \textbf{57.90} \\
    \textbf{MBP w/o masking}        & 90.98 & 69.20 & 70.14 & \underline{6.67} & 40.00 & 26.84 & 73.03 & 26.12 & 76.07 & 54.44 & 75.08 & 55.32 \\
    
    \modelrow{13}{Qwen3-8B-Base}
    Base                            & 88.40 & 65.40 & 65.86 & \textbf{13.33} & 42.50 & 26.84 & 77.79 & \underline{29.46} & 70.82 & 58.48 & 78.24 & 56.10 \\
    SFT                             & 92.95 & 71.60 & 72.54 & 6.67 & 52.50 & 32.72 & 78.21 & 27.90 & 74.80 & \underline{58.56} & 85.09 & 59.41 \\
    Append                          & \underline{93.25} & \textbf{75.40} & \underline{73.00} & \underline{10.00} & \textbf{60.00} & 33.82 & 78.20 & \textbf{35.04} & \textbf{76.45} & 58.49 & 84.17 & \textbf{61.62} \\
    Random                          & 93.03 & 73.80 & 72.26 & 6.67 & 47.50 & 34.56 & \textbf{78.53} & 22.32 & 74.40 & \textbf{58.82} & 84.98 & 58.81 \\
    DIT                             & \textbf{93.93} & 74.40 & \textbf{73.54} & 6.67 & 52.50 & \textbf{35.66} & 78.00 & 26.12 & 73.64 & 58.14 & \textbf{85.53} & \underline{59.83} \\
    \oursrule{13}
    \textbf{MBP}                    & 92.57 & \underline{75.00} & 72.84 & 3.33 & \underline{55.00} & \underline{35.29} & \underline{78.25} & 23.66 & \underline{75.52} & 57.28 & \underline{85.36} & 59.46 \\
    
    \modelrow{13}{Llama3.2-1B-Instruct}
    Base                            & 38.97 & 23.80 & 25.70 & \underline{0.00} & 15.00 & 7.72 & \textbf{42.34} & \underline{27.90} & \textbf{30.63} & \underline{32.43} & \textbf{61.26} & 27.80 \\
    SFT                             & 50.72 & 28.00 & 38.48 & \underline{0.00} & 10.00 & \underline{8.09} & \underline{42.09} & \textbf{28.79} & 28.54 & 31.77 & \underline{58.92} & \underline{29.58} \\
    Append                          & 45.87 & 28.20 & 33.62 & \underline{0.00} & 15.00 & 6.62 & 41.34 & 25.00 & \underline{29.03} & 32.33 & 58.11 & 28.65 \\
    Random                          & \underline{51.93} & 28.60 & 38.94 & \textbf{3.33} & 10.00 & 7.72 & 39.70 & 25.45 & 26.91 & 31.49 & 58.65 & 29.34 \\
    DIT                             & 43.75 & 21.80 & 29.12 & \underline{0.00} & \underline{17.50} & 6.25 & 39.74 & 24.33 & 27.62 & \textbf{33.07} & 56.86 & 27.28 \\
    \oursrule{13}
    \textbf{MBP}                    & 51.63 & \textbf{31.80} & \textbf{41.66} & \underline{0.00} & \textbf{27.50} & \textbf{14.34} & 40.81 & 27.68 & 28.43 & 32.16 & 55.82 & \textbf{31.98} \\
    \textbf{MBP w/o masking}        & \textbf{52.16} & \underline{30.60} & \underline{39.64} & \underline{0.00} & 2.50 & 6.99 & 40.90 & 26.12 & 27.35 & 32.38 & 55.11 & 28.52 \\
    
    \modelrow{13}{Llama3.2-3B-Instruct}
    Base                            & 70.43 & 40.00 & 43.22 & 0.00 & 10.00 & \underline{17.65} & \textbf{59.33} & 27.90 & \textbf{56.28} & \textbf{43.06} & \textbf{76.77} & 40.42 \\
    SFT                             & \underline{76.80} & 46.40 & \underline{55.02} & \underline{3.33} & \underline{27.50} & 13.97 & 55.99 & 28.57 & 49.78 & 41.09 & 73.50 & 42.90 \\
    Append                          & 73.77 & 44.60 & 52.26 & \textbf{6.67} & 25.00 & 15.44 & 56.13 & \underline{29.24} & 50.76 & 40.78 & 74.05 & 42.61 \\
    Random                          & 75.97 & \underline{47.80} & 54.16 & \underline{3.33} & \underline{27.50} & 13.24 & 56.62 & \textbf{29.69} & 51.42 & 41.15 & \underline{74.10} & \underline{43.18} \\
    DIT                             & 70.36 & 38.00 & 43.00 & \underline{3.33} & 17.50 & 12.13 & \underline{57.60} & 28.12 & \underline{52.39} & \underline{42.41} & 74.05 & 39.90 \\
    \oursrule{13}
    \textbf{MBP}                    & \textbf{77.03} & \textbf{48.20} & \textbf{55.92} & \underline{3.33} & \textbf{30.00} & \textbf{19.85} & 56.46 & 25.89 & 50.04 & 41.56 & 73.72 & \textbf{43.82} \\
    \textbf{MBP w/o masking}        & 76.19 & 47.40 & 54.58 & \underline{3.33} & 20.00 & 15.81 & 55.64 & 25.22 & 49.93 & 40.73 & 73.78 & 42.06 \\
    
    \modelrow{13}{Llama3.1-8B-Instruct}
    Base                            & 77.48 & 46.40 & 48.62 & \textbf{6.67} & \underline{26.50} & \underline{22.43} & \textbf{67.56} & 28.79 & \textbf{73.30} & \textbf{49.42} & \textbf{82.05} & \textbf{48.11} \\
    SFT                             & 84.84 & \underline{53.20} & 56.28 & \underline{3.33} & 17.50 & 21.69 & 59.49 & 24.33 & 49.52 & 46.72 & 74.65 & 44.69 \\
    Append                          & 84.46 & 49.80 & 53.08 & \underline{3.33} & 20.00 & 20.59 & 60.84 & 26.56 & 50.24 & \underline{48.26} & 73.67 & 44.62 \\
    Random                          & \underline{85.90} & 52.80 & \underline{56.44} & \underline{3.33} & 20.00 & 20.59 & 59.79 & \underline{29.91} & 49.93 & 46.44 & 73.72 & 45.35 \\
    DIT                             & 77.41 & 44.00 & 47.14 & \underline{3.33} & 20.00 & \textbf{22.79} & 58.51 & \textbf{31.92} & \underline{52.53} & 45.89 & \underline{75.46} & 43.54 \\
    \oursrule{13}
    \textbf{MBP}                    & \textbf{86.28} & \textbf{58.00} & \textbf{61.02} & \textbf{6.67} & \textbf{27.50} & \underline{22.43} & \underline{61.30} & \underline{29.91} & 50.78 & 48.13 & 72.25 & \underline{47.66} \\
    \bottomrule
    \end{tabular}%
    }
    \caption{
    Additional SFT results on mathematical reasoning and general benchmarks.
    All entries report accuracy.
    Average denotes the macro-average over all reported benchmark columns.
    }
    \label{tab:main_math_results}
    \end{table*}

\begin{table*}[t]
    \centering
    \small
    \setlength{\tabcolsep}{6pt}
    \renewcommand{\arraystretch}{1.15}
    \resizebox{\textwidth}{!}{%
    \begin{tabular}{lcccccccccc}
    \toprule
    \textbf{Method}
    & \textbf{MBPP}
    & \textbf{MBPP+}
    & \textbf{HumanEval}
    & \textbf{HumanEval+}
    & \textbf{MMLU}
    & \textbf{GPQA}
    & \textbf{HellaSwag}
    & \textbf{BBH}
    & \textbf{PIQA}
    & \textbf{Average} \\
    
    \modelrow{11}{Qwen3-1.7B-Base}
    Base          & 69.05 & 59.26 & 58.54 & 53.05 & 62.95 & \underline{27.46} & 38.69 & 38.84 & 62.13 & 52.22 \\
    SFT           & 69.58 & \textbf{62.43} & 67.07 & 61.59 & 60.19 & 23.88 & 36.44 & 39.15 & 60.66 & 53.44 \\
    Append        & 69.84 & 60.58 & \underline{70.73} & \textbf{65.24} & 61.53 & \textbf{27.90} & 36.58 & 40.65 & 64.09 & 55.24 \\
    Random        & \underline{70.37} & \underline{61.11} & 67.07 & 62.80 & \textbf{64.03} & 24.78 & \underline{48.36} & 42.50 & \textbf{68.77} & 56.64 \\
    DIT           & 69.84 & 60.58 & \textbf{71.95} & \textbf{65.24} & 63.71 & 25.45 & \textbf{50.40} & \textbf{43.09} & 67.41 & \textbf{57.52} \\
    \oursrule{11}
    \textbf{MBP}         & \textbf{73.02} & \textbf{62.43} & 70.12 & \textbf{65.24} & \underline{63.78} & 25.67 & 44.01 & \underline{42.88} & \underline{68.01} & \underline{57.24} \\

    \modelrow{11}{Qwen3-4B-Base}
    Base          & 74.34 & 64.81 & 76.83 & 72.56 & 73.05 & \underline{30.80} & 63.69 & \underline{56.64} & 73.29 & 65.11 \\
    SFT           & 78.04 & \underline{67.99} & 79.27 & 73.78 & 74.19 & 28.12 & 75.35 & \textbf{56.85} & \underline{78.45} & 68.00 \\
    Append        & \textbf{80.42} & \textbf{68.25} & 80.49 & 74.39 & \underline{74.43} & \textbf{34.15} & \textbf{77.15} & 53.38 & 75.63 & \textbf{68.70} \\
    Random        & 78.84 & 67.20 & \underline{82.32} & \textbf{78.66} & 74.37 & 28.12 & 76.08 & 53.22 & 78.29 & 68.57 \\
    DIT           & 75.93 & 66.93 & 81.10 & 76.22 & 74.36 & 28.12 & \underline{76.38} & 53.43 & 76.77 & 67.69 \\
    \oursrule{11}
    \textbf{MBP}         & \underline{79.10} & 65.08 & \textbf{82.93} & \underline{78.05} & \textbf{74.56} & 29.24 & 74.72 & 54.94 & \textbf{78.89} & \underline{68.61} \\

    \modelrow{11}{Qwen3-8B-Base}
    Base                  & 82.80 & 72.49 & \textbf{88.41} & \textbf{84.15} & 77.79 & \underline{29.46} & 70.82 & 58.48 & 78.24 & 71.40 \\
    SFT                   & \underline{86.51} & \underline{75.13} & 85.98 & 81.71 & 77.96 & 29.24 & 75.33 & 61.25 & 83.30 & 72.93 \\
    Append                & 85.19 & 72.75 & 83.54 & 78.05 & 77.89 & 23.88 & \underline{79.01} & \textbf{61.39} & 82.97 & 71.63 \\
    Random                & 86.24 & \underline{75.13} & 84.76 & 81.10 & \textbf{78.09} & 28.12 & \textbf{79.63} & 59.85 & \textbf{85.75} & \underline{73.19} \\
    DIT                   & 84.92 & 74.07 & \underline{87.20} & 81.10 & 77.75 & \textbf{32.14} & 76.98 & 60.58 & \underline{83.62} & 73.15 \\
    \oursrule{11}
    \textbf{MBP}          & \textbf{87.30} & \textbf{75.66} & 86.59 & \underline{82.32} & \underline{78.02} & 29.24 & 78.16 & \underline{61.31} & 83.03 & \textbf{73.51} \\
    
    \modelrow{11}{Llama3.1-8B-Instruct}
    Base          & 67.46 & 59.52 & 67.68 & 62.20 & \textbf{67.56} & \underline{28.79} & 73.30 & 49.42 & 82.05 & 62.00 \\
    SFT           & 68.25 & 60.05 & 68.29 & \textbf{64.02} & \underline{67.47} & 26.56 & 73.63 & 49.55 & 81.88 & 62.19 \\
    Append        & \underline{69.58} & \textbf{60.85} & 67.68 & 62.20 & 67.44 & 25.45 & \underline{73.87} & 49.47 & \textbf{82.15} & 62.08 \\
    Random        & 69.31 & 60.05 & \underline{68.90} & 63.41 & 67.37 & 25.45 & 73.60 & \textbf{49.76} & \underline{82.10} & 62.22 \\
    DIT           & \textbf{70.11} & \underline{60.58} & 68.29 & \textbf{64.02} & 67.42 & \textbf{29.24} & 73.75 & \underline{49.58} & \underline{82.10} & \textbf{62.79} \\
    \oursrule{11}
    \textbf{MBP}         & 68.52 & 60.05 & \textbf{69.51} & \textbf{64.02} & 67.41 & 28.35 & \textbf{73.93} & 49.36 & \underline{82.10} & \underline{62.58} \\
    \bottomrule

    \end{tabular}%
    }
    \caption{
    Additional SFT results on code generation and general benchmarks.
    Code benchmark entries report pass@1, while general benchmark entries report accuracy.
    Average denotes the macro-average over all reported benchmark columns.
    }
    \label{tab:main_code_results}
    \end{table*}


\begin{table*}[t]
\centering
\small
\setlength{\tabcolsep}{4.5pt}
\renewcommand{\arraystretch}{1.0}
\begin{tabular}{@{}lcccccccc@{}}
\toprule
\textbf{Method} & \textbf{GSM8K} & \textbf{M500} & \textbf{MATH} & \textbf{AIME24} & \textbf{AMC23} & \textbf{Minerva} & \textbf{Avg.} & \textbf{AUC} \\
\modelrow{9}{Qwen3-1.7B-Base}
\textsc{Standard} & 80.52 & 61.20 & 59.68 & 3.33 & 37.50 & 20.96 & 43.87 & 0.366 \\
\oursrule{9}
\textbf{\textsc{MBP}} & \textbf{81.35} & \textbf{61.80} & \textbf{61.24} & 3.33 & \textbf{40.00} & \textbf{21.32} & \textbf{44.84} & \textbf{0.414} \\
\modelrow{9}{Qwen3-4B-Base}
\textsc{Standard} & \textbf{92.19} & 74.40 & 74.30 & 13.33 & 50.00 & \textbf{30.15} & 55.73 & 0.505 \\
\oursrule{9}
\textbf{\textsc{MBP}} & 91.58 & \textbf{76.00} & \textbf{75.26} & \textbf{16.67} & \textbf{57.50} & 23.90 & \textbf{56.82} & \textbf{0.550} \\
\bottomrule
\end{tabular}
\caption{Full GRPO pass@1 accuracy (\%) on six mathematical reasoning benchmarks and normalized training-reward AUC for Qwen3-1.7B-Base and Qwen3-4B-Base. Average denotes the macro-average over the six benchmarks.}
\label{tab:grpo_accuracy_auc_full}
\end{table*}

\end{document}